\documentclass{IEEEtaes}

\usepackage{color, array, amsthm}
\usepackage{graphicx}
\usepackage[colorlinks,urlcolor=blue,linkcolor=blue,citecolor=blue]{hyperref}
\usepackage{amsmath,amsfonts}
\usepackage{algorithmic}
\usepackage{algorithm}
\usepackage[caption=false,font=normalsize,labelfont=sf,textfont=sf]{subfig}
\usepackage{textcomp}
\usepackage{stfloats}
\usepackage{url}
\usepackage{verbatim}
\usepackage{cite}
\usepackage{multicol, multirow}
\usepackage{booktabs}
\usepackage{amsmath}
\usepackage{bbm}
\usepackage{slashed}
\usepackage{amssymb}
\usepackage{eso-pic}
\usepackage{xcolor}

\jvol{XX}
\jnum{XX}
\jmonth{XXXXXXX}
\paper{1234567}
\pubyear{2023}
\doiinfo{TAES.2023.Doi Number}

\begin{document}

\newcolumntype{L}[1]{>{\arraybackslash}m{#1}}
\newcolumntype{C}[1]{>{\centering\arraybackslash}m{#1}}

\title{WAY: Estimation of Vessel Destination in Worldwide AIS Trajectory} 

\author{JIN SOB KIM}
\author{HYUN JOON PARK}
\author{WOOSEOK SHIN}
\affil{Korea University, Seoul 02841, Republic of Korea}

\author{DONGIL PARK}
\affil{SeaVantage, Seoul 06119, Republic of Korea}

\author{SUNG WON HAN}
\affil{Korea University, Seoul 02841, Republic of Korea}

\receiveddate{
This research was supported by Brain Korea 21 FOUR. This research was also supported by Korea Institute for Advancement of Technology(KIAT) grant funded by the Korea Government(MOTIE) (P0008691, The Competency Development Program for Industry Specialist).}

\corresp{{\itshape (Corresponding author: Sung Won Han)}.}

\authoraddress{Jin Sob Kim, Hyun Joon Park, Wooseok Shin, and Sung Won Han are with the School of Industrial and Management Engineering, Korea University, Seoul 02841, Republic of Korea (e-mail: \href{mailto:jinsob@korea.ac.kr}{jinsob}, \href{mailto:winddori2002@korea.ac.kr}{winddori2002}, \href{mailto:wsshin95@korea.ac.kr}{wsshin95}, \href{mailto:swhan@korea.ac.kr}{swhan}; @korea.ac.kr). Dongil Park is with SeaVantage, Seoul 06120, Republic of Korea (e-mail: \href{mailto:dipark@seavantage.com}{dipark@seavantage.com}).}


\markboth{KIM ET AL.}{WAY: ESTIMATION OF VESSEL DESTINATION IN WORLDWIDE AIS TRAJECTORY}
\maketitle

\AddToShipoutPictureBG*{%
  \AtPageLowerLeft{%
    \put(\LenToUnit{0.5\paperwidth}, \LenToUnit{0.8cm}){%
      \makebox(4.7cm,-5.65cm){%
        \colorbox{white}{%
          \begin{minipage}{0.8\paperwidth}
            \centering
            \scriptsize 
            \copyright~2023 IEEE. Personal use of this material is permitted. Permission from IEEE must be obtained for all other republication or redistribution.
          \end{minipage}%
        }%
      }%
    }%
  }%
}

%
\begin{abstract}
The Automatic Identification System (AIS) has recorded near-real-time vessel monitoring data over the years, paving the way for data-driven maritime surveillance methods; concurrently, the data suffer from unrefined, reliability issues and irregular intervals. 
In this paper, we address the problem of vessel destination estimation by exploiting the global-scope AIS data. 
We propose a differentiated data-driven approach recasting a long sequence of port-to-port international vessel trajectories as a nested sequence structure. 
Based on spatial grids, this approach mitigates the spatio-temporal bias of AIS data while preserving the detailed resolution of the original. 
Further, we propose a novel deep learning architecture (WAY) that is designed to effectively process the reformulated trajectory and perform the long-term estimation of the vessel destination ahead of arrival with a horizon of days to weeks. 
WAY comprises a trajectory representation layer and channel-aggregative sequential processing (CASP) blocks. 
The representation layer produces the multi-channel vector sequence output based on each kinematic and non-kinematic feature collected from AIS data. 
Then CASP blocks include multi-headed channel- and self-attention architectures, where each processes aggregation and sequential information delivery respectively.
Then, a task-specialized learning technique, Gradient Dropout (GD), is also suggested for adopting many-to-many training along the trajectory progression on single labels. 
The technique prevents a surge of biased feedback by blocking the gradient flow stochastically using the condition depending on the length of training samples.
Experimental results on 5-year accumulated AIS data demonstrated the superiority of WAY with recasting AIS trajectory compared to conventional spatial grid-based approaches, regardless of the trajectory progression steps. 
Moreover, the data proved that adopting GD in a spatial grid-based approach leads to the performance gain.
In addition, the possibilities of improvement and real-world application with WAY’s expandability in multitask learning for the estimation of ETA was explored.
\end{abstract}
%
\section{INTRODUCTION}
\label{sec:introduction}
S{\scshape ea} routes are the largest market for international goods freight, with the United Nations (UN) stating that more than $80\%$ of the volume of international trade is carried by ship transportation \cite{sirimanne2021review}. 
To secure both the safety and cost efficiency of vessel operations, shipping technologies and infrastructure have made progress toward the advancement of maritime surveillance. 
For example, the Automatic Identification System (AIS), a self-reporting messaging system established on a vessel, is mandatory for international voyages and provides near-real-time surveillance data \cite{SOLAS-5-19}. 
The vast amount of accumulated AIS data opened up possibilities for data-driven analysis of marine traffic patterns, with many recent studies \cite{ristic2008statistical, perera2012maritime, park2021time, gao2018online, nguyen2018multi, forti2020prediction, capobianco2021deep, yu2020deep, zhou2020using, nguyen2021traisformer, nguyen2018vessel} conducted to solve diverse optimization problems in vessel operations.

As an optimization issue, the port congestion problem has arisen in terms of efficient transportation management. 
According to the UN, shortages in port capacities, unreliable schedules, and port congestion, in contrast to the annual growth of international maritime trade keep vessels waiting and have led to a surge in surcharges and fees \cite{sirimanne2021review}.
Knowing the destinations of ongoing ships with an accurate Estimated Time of Arrival (ETA) to ports is the key to resolving these congestion issues in advance. 
AIS data originally contain information on destinations and ETA written by human resources for each data collection. 
However, these reports are unstructured, with unpredictable errors included in the data itself and irregular intervals between each observation, which causes a lack of reliability and challenges the investigation of international marine traffic analysis. Even so, the AIS has recorded a tremendous amount of data since its establishment and contains significant maritime trajectory contexts.

Consequently, large-scale global AIS data rather have been availed the maritime traffic/route analyses and the derivation of nationwide maritime insights \cite{zissis2020distributed, millefiori2021covid}, than exploited in the studies to track a single maritime object.
Meanwhile, for securing safety and auto-navigating systems, many studies \cite{nguyen2018vessel, nguyen2018multi, gao2018online, forti2020prediction, capobianco2021deep, yu2020deep, zhou2020using, nguyen2021traisformer} proposed several advanced data-driven methods to track ongoing vessels at sea and predict their future paths. 
However, these studies were constrained to the Region of Interest (ROI), which is an inevitable condition regarding the sequential length of intercontinental trajectories and to eliminate concerns for unrefined data.

To fulfill the lack of motivation and exploit the AIS data out of the ROI constraint, this study introduces recasting the former data-driven approaches of AIS trajectory sequence representation. Based on the premise that the start and end of the AIS trajectory are known to be a tuple of ports, the departure and the destination, the objective of probability maximization is defined. Thus, to make the premise satisfied, the annotation framework for extracting meaningful segments from the raw AIS sequence is introduced.
Therefore, the verified trajectories comprise AIS messages sharing the same context from departure to destination, underlying the analysis of port-to-port instances. 

Exploiting the annotated trajectory, the deep learning architecture (WAY) is proposed to meet the formularization of a nested sequence structure from the port-to-port AIS trajectory.
To address the problem covering the entire marine region without ROIs, this study further explores the design of the deep learning architecture fitting the diverse properties of AIS data that are multivariate sequences with irregular intervals and non-sequential features. 
The proposed model, WAY, adopts a spatial grid division approach while encompassing the details of the trajectories in a 4-channel vector sequence representation. 
The model utilizes a channel attention mechanism to aggregate the representations at each sequential step and adopts the Transformer-decoder architecture to convey information toward the current step from the past. 
In addition, the task-specialized learning technique, Gradient Dropout, is suggested to temper the surge of feedback bias over the length of individual sequences where the bias occurs due to many-to-many learning for a fixed single destination target according to the trajectory progression.

The remainder of this paper is organized as follows. 
Sec. \ref{sec:related work} provides an overview of prior approaches that have dealt with the refinement of AIS data and the related studies in the field that adopted deep learning methods to process such sequential information. 
Sec. \ref{sec:problem_definition} formulates the problem of destination port estimation given AIS trajectories and the training procedure of the data-driven approach.
Sec. \ref{sec:methodology} presents the details of WAY and applied methods in delivering the long-range estimation. 
Sec. \ref{sec:experiments} demonstrates the leading edge of performance achieved by the proposed method compared to the other related models based on experiments. 
Future studies and conclusions are discussed in Sec. \ref{sec:discussion} and Sec. \ref{sec:conclusion}.
%
\section{RELATED WORK}
\label{sec:related work}
Since AIS messages were first recorded in 2002 \cite{SOLAS-5-19}, a tremendous amount of data has been collected, which contain a wealth of information on international voyages. Currently, the potential of intelligent techniques is in the spotlight, as the affordability of data acquisition increases with advances in storage and processing infrastructure.
By exploiting analyses of massive AIS data, these techniques are expected to enhance or replace conventional Maritime Domain Awareness (MDA) applications, like manual monitoring of traffic patterns at sea, while reducing costs \cite{tetreault2005use, weintrit2009marine, tu2017exploiting}.
MDA appliances, or the awareness of maritime activities in general \cite{tetreault2005use}, have been investigated in recent studies on AIS data with regard to the problem of anomaly detection, route estimation, collision prediction and path planning, amongst others \cite{tu2017exploiting}.
Indeed, the massively accumulated AIS data, continuously collected even today, has proven its potential through large-scale maritime analyses deriving various insights across the globe \cite{millefiori2021covid} as well as modeling worldwide maritime routes \cite{zissis2020distributed}.

However, there are still challenges in applying AIS data to real-world navigation owing to the quality, validity, and accuracy of the data \cite{weintrit2009marine, tu2017exploiting, claramunt2017maritime}.
These challenges arise from errors included in the data, such as the space-temporal mislocation of vessels, missing values, and contextual misinformation.
These are not only caused by machines as sensor/transmission failure, but also by human mistakes such as unstructured/misspelled/unfilled data \cite{kerbiriou2017automatic, harati2007automatic, zhao2018ship, heymann2013plausibility}.
Many state-of-the-art maritime intelligence methods make an effort to overcome these challenges by either preprocessing raw AIS data or considering identification and refinement of the errors as a problem to be addressed.

This section provides a review of related studies regarding methods dealing with AIS data from two perspectives: analyses to extract meaningful context from noisy AIS data, and the former approaches conducted to predict the future location from AIS trajectories.

\noindent \textbf{Cleansing and context extraction of AIS data.}
Vessel-identifying messages are continuously and autonomously transmitted over the AIS system and contain static identification, dynamic position, and future-route indicative information.
On the other hand, in the data provided from recordings, one out of 14 messages includes incorrect/missing identification according to a study that investigated the reliability of AIS data \cite{kerbiriou2017automatic}.
Hence, the data need to be refined and a contextual mining process should be performed before the analysis.

The simplest and most often suggested processing methods were rule-based methods to eliminate errors from raw observations.
Regardless of the task objectives, many AIS-relevant prior studies \cite{mao2018automatic, herrero2019ais, capobianco2021deep, nguyen2021traisformer, sun2020vessel, nguyen2018multi, yu2020deep, zhou2020using} set up boundaries, such as the aforementioned ROI, upon real-valued fields as longitude, latitude, Speed over Ground (SOG), or Course over Ground (COG), to filter noise. The thresholds are set based on either the area of interest, domain knowledge, signal protocol from the AIS system, or considering all of them.
Some priors combined the filtering process with algorithmic methods to deal with the logical integrity of trajectories and improved AIS data quality \cite{heymann2013plausibility, zhao2018ship, sun2020vessel}.
In addition, interpolation of longitudes and latitudes was widely used for preprocessing while handling irregular time-series data \cite{forti2020prediction, capobianco2021deep, yu2020deep}.

Others attempted to detect outliers using statistical approaches and extract the context from the trajectory sequence.
These approaches contributed not only to preprocessing, such as the interpolative substitution of SOG, along with the detection of abnormal velocity peaks, but also to capturing and classifying the behavior of ships.
Clustering-based models were commonly adopted to describe positional anomalies based on time and space perspectives and to discover traffic patterns and meaningful segments of trajectories \cite{capobianco2021deep, mieczynska2021dbscan, pallotta2013vessel, zissis2020distributed}.

\noindent \textbf{AIS trajectory forecasting.}
Defined to predict anticipated future locations given the incomplete trajectory of a moving object, the mobility prediction problem has been considered an important concern for decades in various transport-relevant domains, including urban, maritime, and aviation \cite{georgiou2018moving}.
Formula-based predictions, which rely on motion functions from physics, were one of the earliest approaches in the field \cite{li2003survey}. The constant velocity model, which is one of the well-known linear models in the maritime domain based on the mathematical formula, keeps the SOG and COG constant throughout the prediction \cite{xiao2019traffic}. The control models were investigated in previous studies that adopted a more complex motion formula.
The extended Kalman filter and Ornstein--Ulenbeck (OU) process were employed to track and model vessel dynamic motions at sea \cite{perera2012maritime, millefiori2016long}.
As one of the linear and the non-linear models respectively, these methods are straightforward, simple, and have strength in the robustness of input data quality.
Among the above, in the maritime domain, OU process model based on mean-reverting stochastic processes was adopted to perform long-term prediction of vessel movement over hours in the domain \cite{millefiori2016modeling}.

Recent studies built models based on the history of movements in certain regions by exploiting patterns over the past trajectory of objects.
These data-driven techniques consider not only the movement captured from the target of interest but also of the other observed objects moving in the same area (or the historical movements of the target in the area) \cite{georgiou2018moving}.
Probabilistic approaches based on bridging distributions were proposed, where Bayesian theory was exploited to estimate the probability of endpoints and to infer the intended destination including the future paths from given vessel trajectories, respectively \cite{liang2019destination, ahmad2016bayesian}.
Further, upon the success of deep learning architectures in sequential processing lately, many researchers have conducted studies using deep learning models for AIS data \cite{nguyen2018vessel, nguyen2018multi, gao2018online, forti2020prediction, capobianco2021deep, yu2020deep, zhou2020using, nguyen2021traisformer}.
Representative structures of sequential processing deep neural networks, Recurrent Neural Networks (RNNs) \cite{rumelhart1986learning, werbos1990backpropagation}, Long Short-Term Memory (LSTM) \cite{hochreiter1997long}, and Sequence-to-sequence \cite{sutskever2014sequence} architectures were adopted to capture traffic patterns and estimate the future locations of vessels.
An RNN was proposed to compute the association probability of a new AIS message and an existing track \cite{capobianco2021deep}.
Embedding block of variational RNN encoders was introduced to learn hidden regimes from maritime surveillance tasks \cite{nguyen2018multi}.
The bidirectional LSTM was trained over coordinates to predict the behavior of ships online \cite{gao2018online}. 
Given past AIS trajectory data, an encoder–decoder architecture comprising LSTM generated a predictive coordinate sequence \cite{forti2020prediction, capobianco2021deep}.
Sequence-to-sequence model approach to maritime trajectory prediction was discussed in \cite{capobianco2021deep}, which investigated the intermediate aggregation layer.

Processing trajectories for each AIS message step \cite{forti2020prediction, capobianco2021deep, nguyen2018multi, gao2018online, yu2020deep} has clear limitations in handling global AIS trajectories because of the time-spatial bias within observations that occur from irregular intervals.
Given this bias, many studies either sampled messages in a fixed duration or exploited the linear interpolation of coordinates between the received messages to obtain the trajectory of regular steps \cite{nguyen2018multi, yu2020deep, capobianco2021deep, forti2020prediction}.
However, these methods are difficult to apply to long-range intercontinental trajectories.
Thus, it is not feasible to model the data in such a way, given the number of AIS messages received, which is not related to the physical voyage distance, along with the extreme variances in both length distributions.

Other studies proposed an approach using a token representation that defines vessel positions into a uniformly divided grid area rather than real-valued coordinates \cite{nguyen2018vessel, zhou2020using, nguyen2021traisformer}.
This approach availed Sequence-to-sequence models to be trained on a vessel trajectory as a text sentence and to estimate a future path and destination port, as represented by a sequence of tokens \cite{nguyen2018vessel}.
One attempted to tokenize not only longitude and latitude but also the SOG and COG, and adopted a Transformer–decoder \cite{vaswani2017attention} to deal with AIS trajectory forecasting \cite{nguyen2021traisformer}. 

Based on spatial grid tokens, the bias from irregular intervals can be mitigated by unifying observations within the same spatial area into a single representation \cite{nguyen2018vessel}.
However, the drawback of tokenization is that the approach requires the models to learn the spatial correlation from the bottom, where the correlation can be clearly defined physically by the coordinates.
In addition, by unifying the representation from multiple observations, it is clear that the larger the sampling grid, the more details are lost. By contrast, shrinking the size of the unit space results in the exponential increment of the parameters to learn.
In addition to the tradeoff, it is difficult to specify whether the approach can consider irregular intervals of AIS trajectories because they are not spatial-continuously connected, even in tokenized representations.

In summary, the studies, related to forecast/estimating the future vessel position using AIS trajectory data so far \cite{nguyen2018multi, gao2018online, forti2020prediction, capobianco2021deep, yu2020deep, zhou2020using, nguyen2021traisformer, nguyen2018vessel}, have been constrained under ROI. The constraint should be broken to consider the global scope of analysis, and modeling the port-to-port trajectory at sea should consider various features given real-world AIS data, rather than only regarding kinetic information, which is primarily focused on real-valued coordinates. 

Finally, this article proposes a novel deep-learning approach for the task, estimation of the port destination given the vessel operation trajectory from its departure, exploiting the worldwide AIS data. The contribution of this work focuses on the followings:
\subsubsection*{\bf Contribution}
\begin{itemize}
\item{\bf Recasting data-driven approach of AIS trajectory} \\
\text{\space\space\space} AIS data have been dealt with as a sequence of either message-wise observations or spatial grid representation units. However, neither of them fits to estimate the long-range destination of international vessel operations far ahead days to even a few months of the arrival due to each reason; the irregular intervals form spatio-temporal bias and the coarse portrayal loses the details, respectively. This work suggests recasting the problem of global port destination estimation with a novel approach based on spatial grids mitigating the spatio-temporal bias while preserving the detailed representation.
\item{\bf Establishing a novel deep learning architecture} \\ 
\text{\space\space\space} This work also includes a novel framework of deep learning model that meets with the setup of problem reformulated. the model comprises advanced deep-learning architectures adopting multi-channel representation, multi-headed self- and channel- attention techniques.
Compared to the conventional methods of vessel trajectory forecasting, the proposed model fully utilizes the given AIS information, not only the kinematic features but also the non-kinetic context such as ship type and departure port, and estimates the port destination.
\item{\bf Introducing a task-specialized learning technique} \\
\text{\space\space\space} As the problem leads the model training to maximize the probability of the target destination given an incomplete trajectory, a surge of feedback may occur in the training procedure according to the varying length of vessel motions. Here, a task-specialized learning technique Gradient Dropout prevents the model to be biased in such a many-to-many learning process by stochastically suppressing the surge depending on the length of training samples.
\item{\bf Comparing different deep learning methods} \\
\text{\space\space\space} The proposed method is compared to the conventional spatial grid-based approach with different deep learning benchmark algorithms. Performing the destination estimation task, the experiment shows the robustness of the proposed model dealing with reformulated representation against the other prior approaches. In addition, further experiment depicts the model in terms of feature availability, inter-channel processing methods, and efficiency.
\end{itemize}
%
\section{PROBLEM DEFINITION}
\label{sec:problem_definition}
In this section, the problem of port destination estimation is formalized. Assuming any vessel operation proceeds between a certain tuple of departure and destination ports, AIS trajectory data can be represented as a set of $\{\mathcal{P}_{\alpha}, \mathcal{P}_{\omega}, \mathcal{S}_{T} \}$. The symbols $\alpha$ and $\omega$ denote the departure and destination among the total set of ports $\mathcal{P}$ respectively, and $\mathcal{S}_{T}$ is the time-ordered sequence of $T$ AIS messages received along the space-time progression of the operation. To address the problem of destination estimation ahead of vessel arrival, the objective of learning from a data-driven approach is to maximize the conditional probability of the destination port, given the departure and the ordered sequence of AIS messages to the current time $t$.
\begin{equation}
\label{eq:trajectory_definement_original}
\begin{aligned}
\begin{split}
\mathcal{S}_{T} = &\{x_{t}; t=1, ..., T \} \\
\mathrm{maximize}&\text{\space}p(\mathcal{P}_{\omega} | x_{1},..., x_{t}, \mathcal{P}_{\alpha})
\end{split}
\end{aligned}
\end{equation}

Given that the preferable condition of worldwide data collection, such as the sequence of uniform intervals thoroughly containing flawless messages, the estimation can be ideally processed by model function $\mathcal{M}$ with parameter $\theta$. Hence the formulation of the destination estimation of the given trajectory at step $t$, and the training procedure can be defined as:
\begin{equation}
\label{eq:model_estimation_original}
\begin{aligned}
\begin{split}
\hat{\mathcal{P}_{\omega}} &= \mathrm{argmax}_{\mathcal{P}}\mathcal{M}(x_{1:t}, \mathcal{P}_{\alpha}; \theta) \\
\hat{\theta} &= \mathrm{argmin}_{\theta}\frac{1}{N}\sum^{N}_{i=1}{\mathcal{L}(\mathcal{M}(x_{1:t}^{i}, \mathcal{P}_{\alpha}^{i};\theta), \mathcal{P}_{\omega}^{i})}
\end{split}
\end{aligned}
\end{equation}
where the procedure aims to minimize the error of loss function $\mathcal{L}$ from given $N$ training trajectory samples, $x_{1:t}^{i}$, $\mathcal{P}_{\alpha}^{i}$, and $\mathcal{P}_{\omega}^{i}$ are the input AIS sequence, departure port, and target destination, correspondingly.

However, the formulation is difficult to be directly applied to real-world AIS data for two main reasons.
First, real-world AIS data has been collected forming irregular intervals. The properties cause spatio-temporal bias about data existence and affect the training to overfit certain excessively data-populated regions.
Second, AIS data itself is the cumulative result of a tremendous number of near-real-time records with unverified transmission errors included, while the data is not separately annotated as port-to-port operations.
Next two subsections present the mitigation of the issues above.
%
\subsection{Spatial Grid Unit Processing}
\label{subsec:formulation_grid_unit_processing}
As discussed in Sec. \ref{sec:related work}, a prior spatial grid token-based processing approach has been proposed to resolve the spatio-temporal bias issue \cite{nguyen2018vessel}. 
Adopting the token-based approach, given $\mathcal{S}_{K}$ reconstructs the original sequence $\mathcal{S}_{T}$, the objective formulations can be recast as below:
\begin{equation}
\label{eq:trajectory_definement_gridunit}
\begin{aligned}
\begin{split}
\mathcal{S}_{T} \simeq \mathcal{S}_{K} = \{g_{k}; k=&1, ..., K \text{\space} | \text{\space} g_{k} \leftarrow (x_{i},...,x_{i+j}) \} \\
\mathrm{maximize}&\text{\space}p(\mathcal{P}_{\omega} | g_{1},..., g_{k}, \mathcal{P}_{\alpha})
\end{split}
\end{aligned}
\end{equation}
where $\mathcal{S}_{K}$ comprises the element $g_{k}$ which is the spatially integrated representation of a non-overlapping subset of $\mathcal{S}_{T}$ each, that is in continuous time order from $i$ to $i+j$. 
Accordingly, the training and estimation of the model would be recast as below, where the unit step of trajectory processing, $g_{k}$ replaces the $x_{t}$.
\begin{equation}
\label{eq:model_estimation_gridunit}
\begin{aligned}
\begin{split}
\hat{\mathcal{P}_{\omega}} &= \mathrm{argmax}_{\mathcal{P}}\mathcal{M}(g_{1:k}, \mathcal{P}_{\alpha}; \theta) \\
\hat{\theta} &= \mathrm{argmin}_{\theta}\frac{1}{N}\sum^{N}_{i=1}{\mathcal{L}(\mathcal{M}(g_{1:k}^{i}, \mathcal{P}_{\alpha}^{i};\theta), \mathcal{P}_{\omega}^{i})}
\end{split}
\end{aligned}
\end{equation}
On the other side, however, the coarsely represented sequence $\mathcal{S}_{K}$ loses the locally detailed navigational pattern information from the original.

Resolving the tradeoff, the proposed method in this paper adopts rearranging the original trajectory as a nested sequence structure as follows:
\begin{equation}
\label{eq:trajectory_definement_final}
\begin{aligned}
\begin{split}
\mathcal{S}_{T} = \mathcal{S}_{K} = \{G_{k}&=\{x_{i},...,x_{i+j};g_{k}\}; k=1, ..., K \} \\
\mathrm{maximize}&\text{\space}p(\mathcal{P}_{\omega} | x_{1},..., x_{t}, \mathcal{P}_{\alpha}) \\ 
\Leftrightarrow \text{\space} &\mathrm{maximize}\text{\space}p(\mathcal{P}_{\omega} | G_{1},..., G_{k,t}, \mathcal{P}_{\alpha})
\end{split}
\end{aligned}
\end{equation}
where $G_{k}$ encompasses not only the spatial representation of the grid unit $g_{k}$ integrating the continuous subset $\{x_{i},...,x_{i+j}\}$ from the original trajectory $\mathcal{S}_{T}$ but also the subset itself. Since, $G_{k,t}$, the last given element of $\mathcal{S}_{K}$ on the current step $t=i+j$ of the original sequence $S_{T}$, has a state dependency on $t$, the maximization objective of the conditional probability from the given rearranged trajectory meets that of the original.

Finally, the formulation of destination estimation and of the training can be recast as:
\begin{equation}
\label{eq:model_estimation_final}
\begin{aligned}
\begin{split}
\hat{\mathcal{P}_{\omega}} &= \mathrm{argmax}_{\mathcal{P}}\mathcal{M}(G_{1:k,t}, \mathcal{P}_{\alpha}; \theta) \\
\hat{\theta} &= \mathrm{argmin}_{\theta}\frac{1}{N}\sum^{N}_{i=1}{\mathcal{L}(\mathcal{M}(G_{1:k,t}^{i}, \mathcal{P}_{\alpha}^{i};\theta), \mathcal{P}_{\omega}^{i})}
\end{split}
\end{aligned}
\end{equation}
where the input sequence $G_{1:k,t}^{i}$ comprises an ordered sequence of the non-overlapping subsequences of the original $S_{t}^{i}$ and the sequence of spatially integrated unit grid representations respectively. Accordingly, the model can process the trajectory data over unit steps upon the spatially uniform grid without losing the locally detailed information of vessel operations.
\begin{figure*}[!t]
\centering
\includegraphics[scale=0.60]{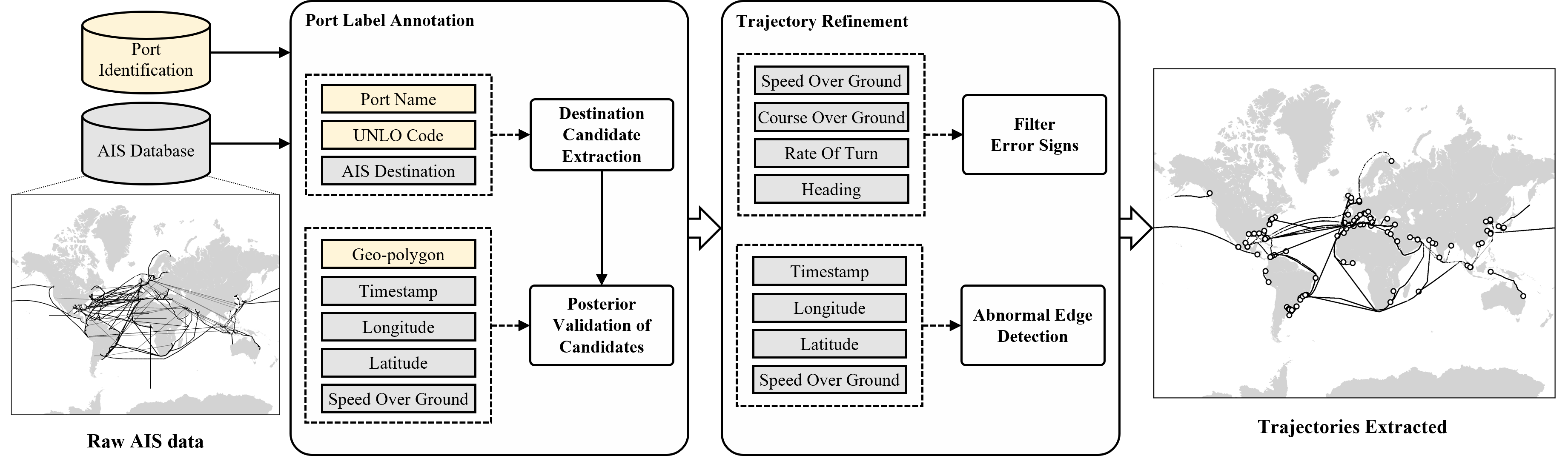}
\caption{Overall annotation processing framework.}
\label{fig:annotation_framework_overfiew}
\end{figure*}
%
\subsection{Annotation Framework}
\label{subsec:annotation_framework}
To meet the needs of the assumption of the data-driven formulation indeed, raw AIS data need to be annotated and refined as a set of meaningful segments that the start and end of each vessel operation are defined. The segment, which can be called port-to-port operation, is the trajectory recorded from a set of continuous AIS observations that share the same departure and destination. The refinement process is the verification and removal of the error that occurred during a given port-to-port trajectory, which causes the illogical shift of vessel movement.

In this work, the port identification dataset and the AIS data are coupled to perform the segmentation and refinement of raw AIS data.
The segmentation of port-to-port trajectory is conducted along the features from the AIS messages received, by exploiting the Damerau--Levenshtein distance \cite{levenshtein1966binary, damerau1964technique} and port-wise boundary thresholding, where the detailed implementation is described in Appendix \ref{appendix:port_label_annotation}.
Afterward, the segmented port-to-port trajectories undergo the refinement process with the clustering algorithm DBSCAN, of detecting abnormal shifts.
The detail of Implementing DBSCAN as verifying the erroneous transmission is depicted in Appendix \ref{appendix:trajectory_refinement}.
The overall procedure of the data annotating framework is shown in Fig. \ref{fig:annotation_framework_overfiew}, which illustrates the refined result of 5-year accumulated AIS data from a single vessel.
%
\section{METHODOLOGY}
\label{sec:methodology}
\begin{figure*}[!t]
\centering
\includegraphics[scale=0.8]{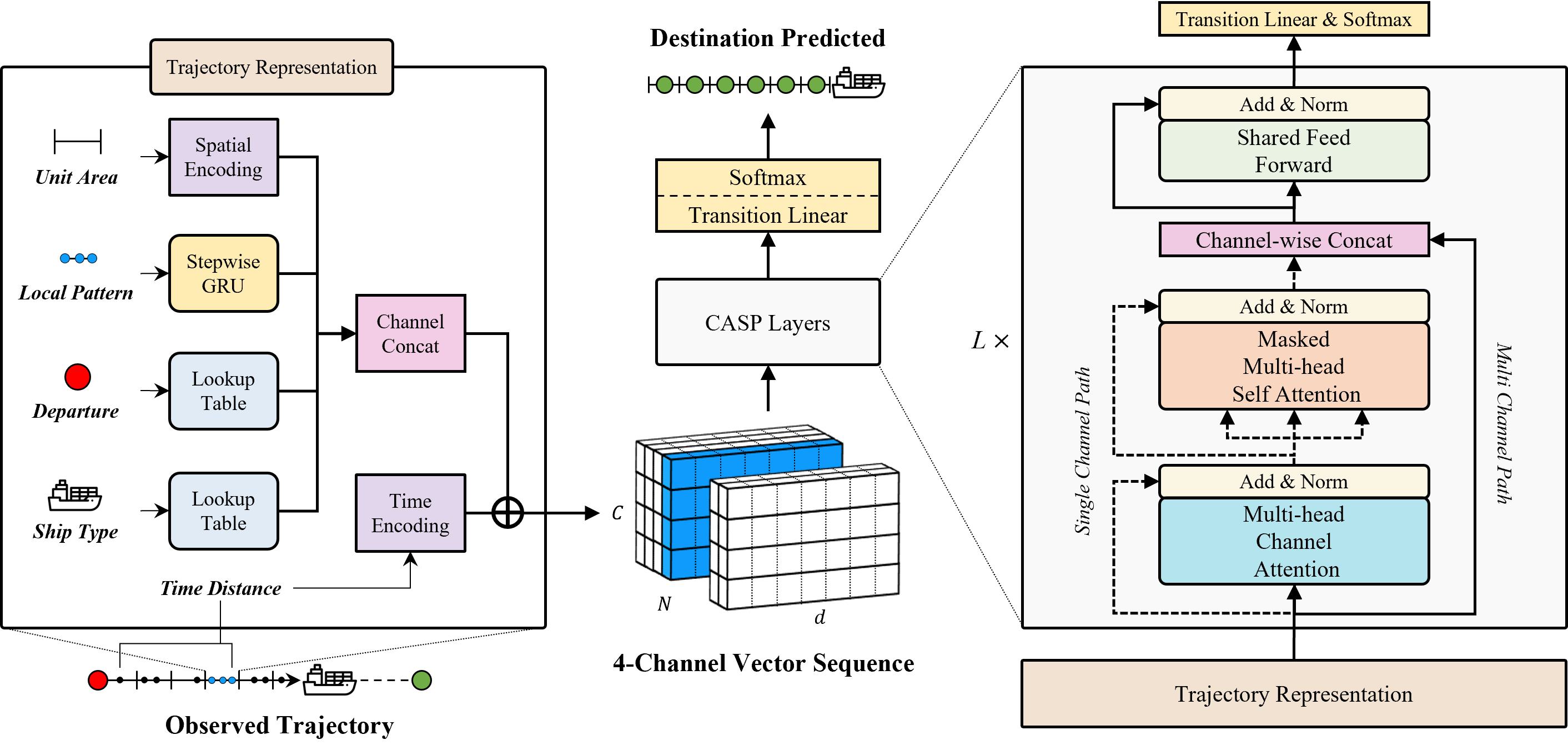}
\caption{Overview of WAY architecture.}
\label{fig:model_overview}
\end{figure*}

This section presents WAY, the architecture for processing AIS data, with a novel approach toward global trajectory representations.
In addition, Gradient Dropout learning technique is proposed to mitigate the feedback bias issue proportional to the data length while designing a many-to-many training framework.
This technique is only applied to the training set and detached for inference.
The overall framework of the proposed method is illustrated in Fig. \ref{fig:model_overview}, where $C$ denotes channels, $N$ implies the number of grid areas observed, $d$ is the hidden dimension size of the model, and $L$ indicates the number of layers repeated as a stack, and WAY is shown to process the trajectories based on 4-channel vector sequences.
\begin{figure}[!t]
\centering
\includegraphics[scale=0.78]{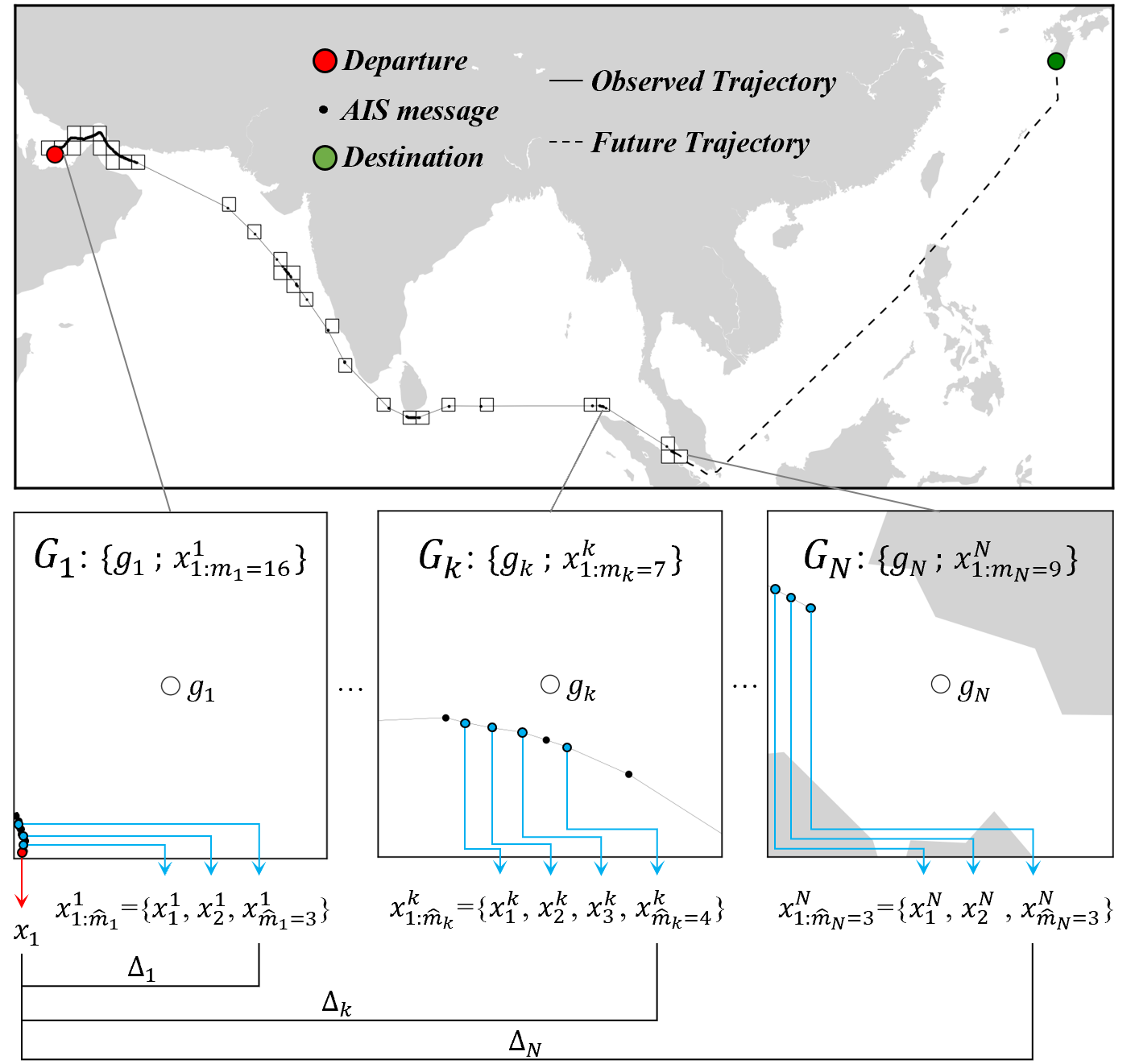}
\caption{An example of reorganizing AIS trajectory in WAY. The \textcolor{red}{Red} arrow indicates the very first observation in given trajectory, and the \textcolor{blue}{Blue} arrows show the sampled observations defining subsequences from each region. $\Delta$ denotes the time distance measured in days.}
\label{fig:trajectory_representation}
\end{figure}
%
\subsection{Trajectory Representation}
\label{subsec:trajectory_representation}
An observed trajectory is processed within each unit area in the WAY architecture.
To do so, the raw observations are reconstructed and handled by nature before the main process began.
Let us denote $x_{1:T} = \{x_{1}, \dots ,x_{T} \} \in \mathbb{R}^{T \times f} $ as a sequence of AIS observations from an ongoing ship at time step $T$, where $f$ indicates the number of AIS fields used as features that contain timestamps, longitudes, latitudes, etc.
The given sequence can therefore be reorganized using uniform spatial grids to $\{G_{1}, \dots ,G_{N} \}$, where each grid element $G_{k} = \{g_{k} ; x^{k}_{1:m_{k}} \}$ for $k \in \{1, \dots , N \}$ includes the identity coordinate $g_{k}$ and a subset of observations inside $x^{k}_{1:m_{k}} = \{x_{j+1}, \dots ,x_{j+m_{k}}\} \subset x_{1:T}$, and where $j=\sum_{i=1}^{k-1} m_{i}$ and $T=\sum_{i=1}^{N} m_{i}$. 

Thereafter, samples of $\hat{m}_{k}$ observations are taken dynamically from $x^{k}_{1:m_{k}}$, where $\hat{m}_{k}$ follow a Poisson distribution with the parameter $\lambda=5$.
The intention of sampling is to reflect the circumstances under irregular intervals and consider any moment of trajectory progression when and wherever the observation possibly exists in regions.
In addition, sampling can moderate the flooding of redundant AIS records generated by such intervals and also by the reason owing to the timing issue between transmission and receiving.
Thus, a subsequential trajectory $x^{k}_{1:m_{k}\gets\hat{m}_{k}}$ is redefined by the sampled observations in each $G_{k}$.

An example of this rearrangement is visualized in Fig. \ref{fig:trajectory_representation}, while the longitude, latitude, and timestamp fields included in $x^{k}_{1:m_{k}}$ are transformed into a manner of relativeness to each grid area belonged.
Based on each grid center coordinate $g_{k}=(\lambda, \phi)$, the observed longitudes and latitudes $(\lambda_{i}, \phi_{i}) \in x^{k}_{i}$ for $i \in \{1,...,m_{k}\}$ are converted into $(\lambda-\lambda_{i}, \phi-\phi_{i})$, and the $m_{k}$ timestamps are converted into $m_{k}$ time distances, measuring the progression in days from the $0$ distance for the first sample $x^{k}_{1}$.
The trajectory contains the non-sequential features as well, acknowledged from the annotation and the ship itself, which are the departure port $Y_{x}$ and vessel type $S_{x}$, respectively.
Subsequently, the organized model input $x = \{G_{1:N} ; Y_{x} ; S_{x} \}$ is handled in the representation layer by each feature component, as described below:
\subsubsection*{\bf Description of AIS trajectory components}
\begin{itemize}
\item{\bf Unit Area.} A sequence of center coordinate $(\lambda, \phi)$ tuples of the spatial grid from the observed, denoted by $\{ g_{1}, ... ,g_{N} \} \in \mathbb{R}^{N \times (\lambda, \phi)}$.
\item{\bf Local Pattern.} A sequence of subsets from the observed; each includes kinetic patterns corresponding to the unit area. The sequence is denoted by $\{ x^{1}_{1:m_{1}}, \dots ,x^{N}_{1:m_{N}} \} \in \mathbb{R}^{N \times M \times f}$, where $M$ denotes a set of subsequential lengths.
\item{\bf Departure.} A categorical value of ports $Y_{x} \in \mathbb{N}$.
\item{\bf Ship Type.} A categorical value of ships $S_{x} \in \mathbb{N}$.
\item{\bf Time Distance.} A sequence of time intervals between each last observation in subsets and the very first from the trajectory, denoted by $\{ \Delta_{1}, \dots ,\Delta_{N} \} \in \mathbb{R}^{N}$ where $\Delta_{N} = \text{time}_{\Delta}(x^{N}_{m_{N}} - x_{1})$ and the intervals are measured in days.
\end{itemize}
%
\subsubsection{Spatial Encoding}
\label{subsubsec:spatial_encoding}
\begin{figure}[!t]
\centering
\includegraphics[scale=0.55]{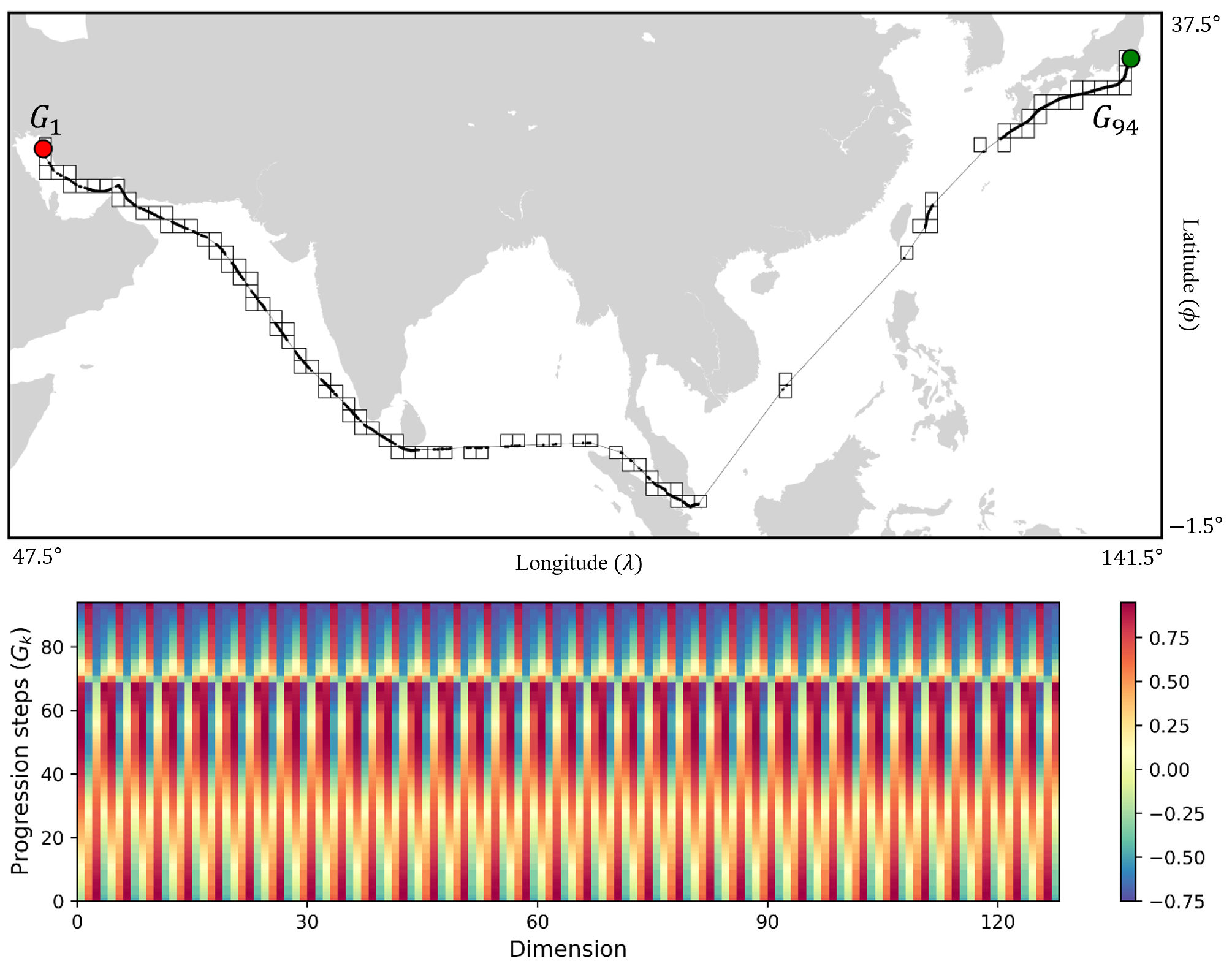}
\caption{The example output of Spatial Encoding, given a sequence of grid coordinates from the above trajectory.}
\label{fig:spatial_encoding_sample}
\end{figure}
\begin{figure}[!t]
\centering
\includegraphics[scale=0.52]{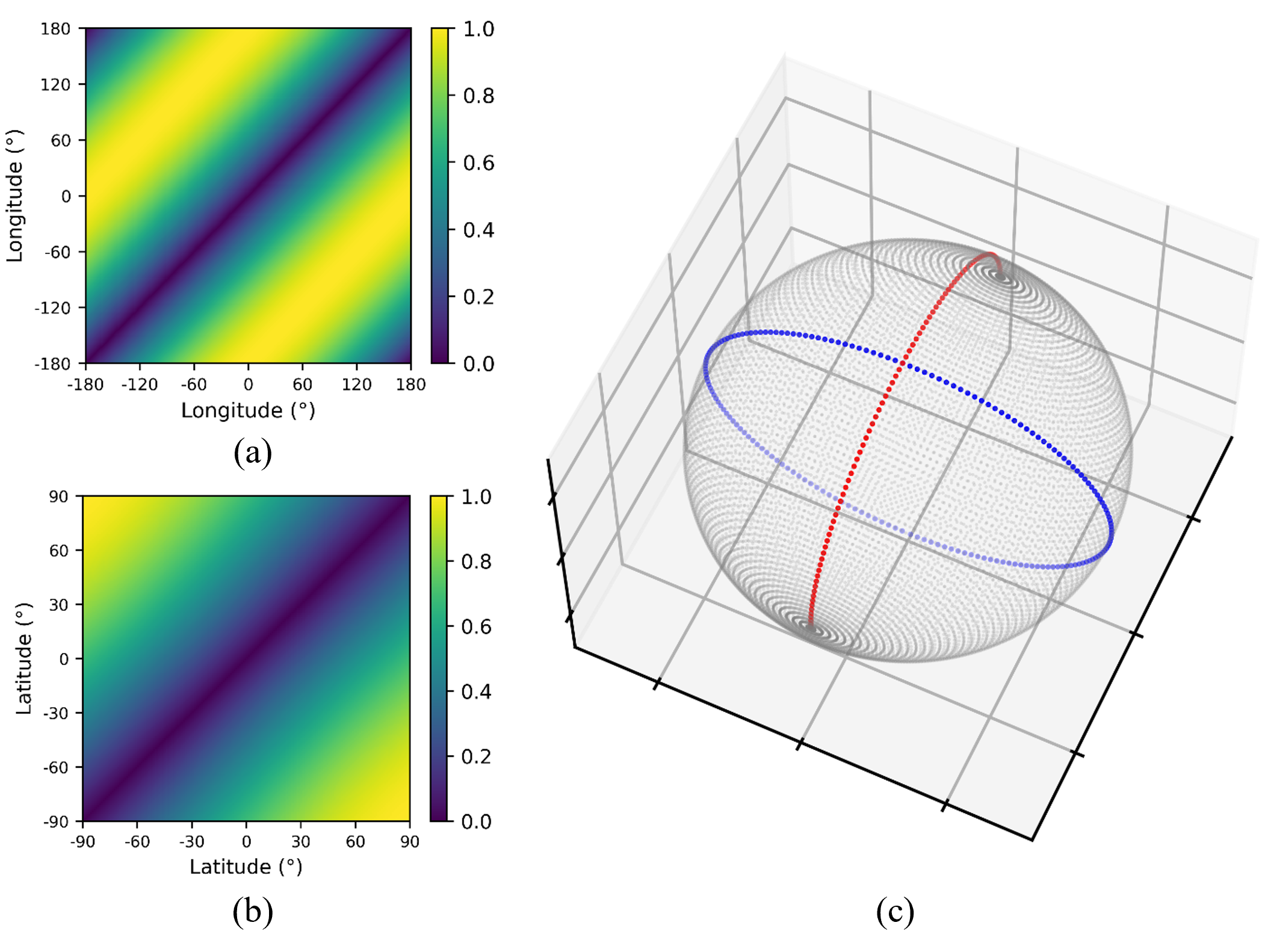}
\caption{An example of Spatial Encoding result. (a) indicates the longitude-wise Euclidean distances between encoded vectors, (b) is the distance between vectors arranged in latitudes, and (c) visualizes the 3-dimensional PCA result of encoded vectors $\in \mathbb{R}^{d}$ from the whole {$1\times1$} (${}^{{\circ}^{2}}$) grids, where the \textcolor{blue}{Blue} are sampled in (a) and \textcolor{red}{Red} as (b).}
\label{fig:spatial_encoding}
\end{figure}

Inspired by \emph{positional encoding} from Transformer \cite{vaswani2017attention}, Spatial Encoding (SE) module is designed to transform real-valued coordinates into high-dimensional vectors.
Given an $N$ observed sequence of the unit area coordinates $x \in \mathbb{R}^{N \times (\lambda, \phi)}$ and hidden size $d$, the module produces the output vector sequence $x \in \mathbb{R}^{N \times d}$ as follows:
\begin{equation}
\label{eq:spatial_encoding}
\begin{aligned}
\begin{split}
\mathrm{SE}_{(\lambda, \phi, 4i)} \text{\space\space\space} &= \cos(\mathrm{g}(\phi, 4i)) \cdot \sin(\mathrm{g}(\lambda, 4i)) \\
\mathrm{SE}_{(\lambda, \phi, 4i+1)} &= (\log{\pi})^{2} \cdot \sin(\mathrm{g}(\phi, 4i))  \\   
\mathrm{SE}_{(\lambda, \phi, 4i+2)} &= \cos(\mathrm{g}(\phi, 4i)) \cdot \cos(\mathrm{g}(\lambda, 4i)) \\
\mathrm{SE}_{(\lambda, \phi, 4i+3)} &= -(\log{\pi})^{2} \cdot \sin(\mathrm{g}(\phi, 4i))      
\end{split}
\end{aligned}
\end{equation}
where the function $\mathrm{g}(q, 4i) = q / (2\pi)^{4i/d^{2}}$ operates with radian unit $(\lambda, \phi)$, and $i$ specifies the dimension.
The combinations of the sinusoidal functions are motivated by the conversion definition of polar to Cartesian coordinate system, and the frequency from the division term of function $g(q, 4i)$ is set to $2\pi$. Therefore, the produced vector comprises the sinusoidal wave of marginally different frequencies, where each $4i^{th}$ and $4i+2^{th}$ dimensions are based on the wavelengths from $\pi$ to $\pi\cdot(2\pi)^{-1/d}$, and $4i+1^{th}$ and $4i+3^{th}$ dimensions corresponds to the wavelengths from $2\pi$ to $(2\pi)^{1-1/d}$.
Fig. \ref{fig:spatial_encoding_sample} illustrates the actual output of the SE module with $d=128$, given the sequence of the grid coordinates from the example trajectory.
Fig. \ref{fig:spatial_encoding}a and Fig. \ref{fig:spatial_encoding}b show the scaled distance distributions between the encoded vectors from sets of each longitude- and latitude-wise sampled coordinates, respectively.
As the transformed representations have the cyclic distance with the longitudes and represent the diverging distance distribution with the latitudes, the set of outputs from SE module preserves the spatial distance concept of the original coordinates.
As a result, the encoded vector from each grid coordinate represents its own positional identity on the spherical surface, which can be explained by the dimensionality reduction of Principal Component Analysis (PCA) \cite{pearson1901liii} (see Fig. \ref{fig:spatial_encoding}c).
%
\subsubsection{Step-wise Gated Recurrent Unit}
\label{subsubsec:stepwise_gru}
To capture the ship's behavioral details and specify local patterns in uniform grid areas, Gated Recurrent Unit (GRU) \cite{cho2014learning} is used to process each subsequence, which employs the concept of LSTM, updating hidden and cell states via gate functions while using fewer parameters.
Given the individual subsequence $x^{k} \in \mathbb{R}^{m_{k} \times f}$, the state update at the current step $t \in [1, \dots ,m_{k}]$ is defined by the input $x^{k}_{t}$, sigmoid function $\sigma$, update gate $u_{t}$, and reset gate $r_{t}$, as follows:
\begin{equation}
\label{eq:gru_state_update}
\begin{aligned}
u_{t} &= \sigma(W_{u}x^{k}_{t} + U_{u}h_{t-1}+b_{u}) \\ 
r_{t} &= \sigma(W_{r}x^{k}_{t} + U_{r}h_{t-1}+b_{r}) \\
\Tilde{h}_{t} &= \tanh(W_{h}x^{k}_{t}+(r_{t} \odot h_{t-1}) U_{h} + b_{h}) \\
h_{t} &= (1-u_{t}) \odot h_{t-1} + u_{t} \odot \Tilde{h}_{t}
\end{aligned}
\end{equation}

Batchifying the process among the $N$ sequence steps, each input subsequence $x^{k} \in \mathbb{R}^{m_{k} \times f}$ yields $h_{m_{k}} \in \mathbb{R}^{d}$ using the ($L/2$)-layered GRU stack. Consequently, the set of last hidden states $[h_{m_{1}}; \dots ;h_{m_{N}}] \in \mathbb{R}^{N \times d}$ represents the sequence of local navigational patterns.
%
\subsubsection{Semantic Representations and Time Encoding}
\label{subsubsec:semantics_and_time_encoding}
Objectifying a port-to-port trajectory as a single analytic instance, departure and vessel types contain a semantic relationship toward the port destination.
That is, international shipments do not occur randomly, but are involved in national trading relations, goods, and needs.
Learnable parameters are set for $Y$ worldwide ports and $S$ types of ships such that each lookup table can be described by $W^{Y} \in \mathbb{R}^{Y \times d}$ and $W^{S} \in \mathbb{R}^{S \times d}$, respectively.
The tables return the vectors of $W^{Y}_{x} \in \mathbb{R}^{d}$ and $W^{S}_{x} \in \mathbb{R}^{d}$ indexed by departure $Y_{x}$ and ship type $S_{x}$.

Thereafter, the vectors are sequentialised by adding time information in order. 
Modifying \emph{positional encoding} \cite{vaswani2017attention}, Time Encoding (TE) module generates representations of the temporal progression of a given trajectory.
Whereas the fixed uniform-discrete offset is given to the original encoding procedure, AIS data comprises the sequence of irregular intervals.
Therefore the procedure has to be adjusted to accept a real-value offset input $\Delta$, the day-unit time distance, and operate as a continuous function.
Based on the timestamps of each sampled subsequence as shown in Fig. \ref{fig:trajectory_representation}, given the time distances between the last samples and the first observation of whole trajectory, the equation to process input $x \in \mathbb{R}^{N}$ can be defined as:
\begin{equation}
\label{eq:time_encoding}
\begin{aligned}
\begin{split}
\mathrm{TE}_{(\Delta, 2i)}\text{\space\space\space}&=\cos(\Delta/1000^{2i/d}) \\
\mathrm{TE}_{(\Delta, 2i+1)}&=\sin(\Delta/1000^{2i/d})
\end{split}
\end{aligned}
\end{equation}
where $i$ specifies the dimension.
To sufficiently cover the maximum temporal progression of the vessel operations and consider the adequate intervals within the encoded representations, the frequency of the sinusoidal functions in TE is set to 1000. 
Subsequently, the TE output is produced as $x \in \mathbb{R}^{N \times d}$, forming a geometric progression in each dimension based on wavelengths from $2\pi$ to $1000\cdot2\pi$.
The sequence of time-encoded vectors is added to $W^{Y}_{x}$ and $W^{S}_{x}$ each while duplicating their identical for $N$ steps, as well as to the sequence outputs that were defined from Sec. \ref{sec:methodology}\ref{subsec:trajectory_representation}\ref{subsubsec:spatial_encoding} and Sec. \ref{sec:methodology}\ref{subsec:trajectory_representation}\ref{subsubsec:stepwise_gru} each.
Accordingly, the concatenation of such representations yields a 4-channel vector sequence $x \in \mathbb{R}^{C \times N \times d}$, where $C=4$.
An observed AIS trajectory is time-sequentially depicted by perspectives of global space, local patterns, and semantics in the respective channels.
%
\subsection{Channel Aggregative Sequence Processing}
\label{subsec:channel_aggregative_sequence_processing}
Based on the multi-channel vector sequence output from the representation layer, Channel Aggregative Sequence Processing (CASP) layer is proposed to aggregate each distinct nature of AIS data distributed over channels and convey the sequential information toward the current step of the decision.
CASP layer comprises multi-head channel attention (MCA), masked multi-head self-attention (MSA), and shared feed-forward (SFF) network modules responsible for current feature aggregation, sequential processing, and decoding, respectively.
%
\subsubsection{Multi-head Channel Attention}
\label{subsubsec:multihead_channel_attention}
\begin{figure}[!t]
\centering
\includegraphics[scale=0.72]{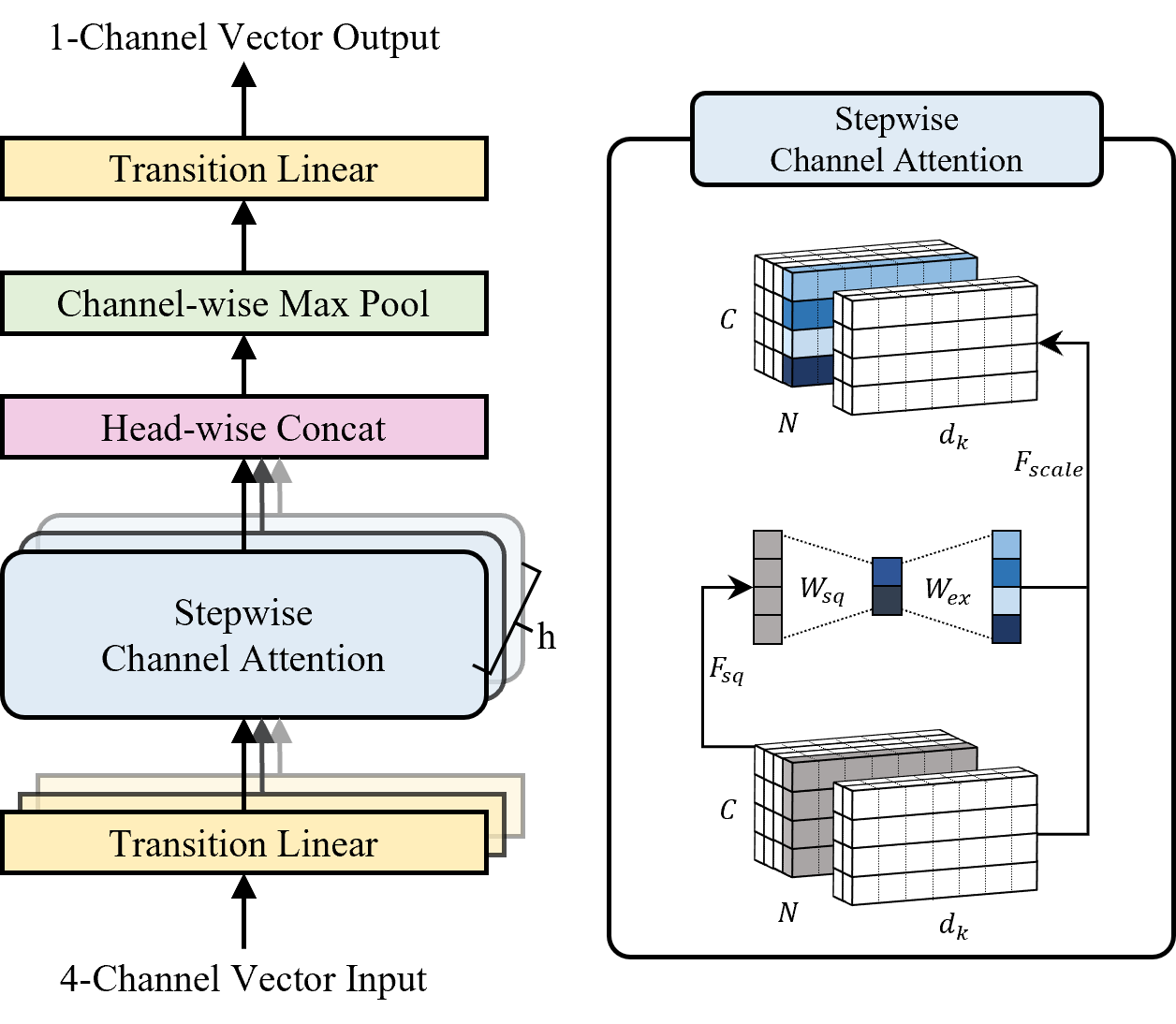}
\caption{Multi-head Channel Attention Module.}
\label{fig:multihead_channel_attention}
\end{figure}
Channel Attention was first proposed in the Squeeze-and-Excitation network \cite{hu2018squeeze} in the computer vision domain. This method involves conducting an information comparison among channels and emphasizing each by the compared results.
The concept is extended in WAY to a multi-perspective comparison by applying an identical set of linear transitions to channels, as shown in Fig. \ref{fig:multihead_channel_attention}.
MCA aggregates the current step representations from each channel into a single vector comprising accentuated dimensions.
From individual steps $x \in \mathbb{R}^{C \times d}$, the perspectives are spread via $W^{tr}_{i} \in \mathbb{R}^{d \times d_{k}}$ for $i \in [1, \dots ,h]$, where $h$ denotes the number of heads and $d_{k}$ is dimensions per head.
Thus, the current step $x \in \mathbb{R}^{h \times C \times d_{k}}$ is given, and the equation of multi-headed attention is as follows:
\begin{equation}
\label{eq:multihead_channel_attention}
\begin{aligned}
\begin{split}
[z_{avg};z_{max}] &= z = [\mathrm{F}_{sq}^{\text{AVG}}(x); \mathrm{F}_{sq}^{\text{MAX}}(x)] \\
[z'_{avg};z'_{max}] &= z' = \text{max}(0, zW_{sq})W_{ex}  \\
\alpha_{C} &= \sigma(z'_{avg} + z'_{max}) \\
\Tilde{x} &= \alpha_{C} \odot x
\end{split}
\end{aligned}
\end{equation}
where $\mathrm{F}_{sq}^{\text{AVG}}$ is the average pooling function and $\mathrm{F}_{sq}^{\text{AVG}}$ is the max pooling in terms of dimensions in each channel.
The squeezed $z_{avg}, z_{max} \in \mathbb{R}^{h \times C}$ is transformed into the corresponding $z'_{avg}, z'_{max} \in \mathbb{R}^{h \times C}$ using the shared parameters $W_{sq}$ and $W_{ex}$.
The procedures for each head are performed in parallel using $W_{sq} \in \mathbb{R}^{h \times C \times C/\gamma}$ and $W_{ex} \in \mathbb{R}^{h \times C/\gamma \times C}$, where $\gamma=2$ and ReLU \cite{nair2010rectified} activation is applied to the bottleneck.
The attention weight $\alpha_{C} \in \mathbb{R}^{h \times C}$ emphasizes the channels of the input $x$, where the sigmoid function $\sigma$ contracts the sum of the intensities in the range of $(0, 1)$.
The emphasized tensor $\Tilde{x} \in \mathbb{R}^{h \times C \times d_{k}}$ are transposed to $\Tilde{x} \in \mathbb{R}^{C \times (h \cdot d_{k})}$ by head-wise concatenation and then a max pooling function is applied over channels, extracting the most highlighted values for each dimension.
Thereby 1-channel vector output $\Tilde{x} \in \mathbb{R}^{(h \cdot d_{k})}$ is provided, and the final result of the module $\Tilde{x} \in \mathbb{R}^{d}$ is produced by the linear transition of $W^{out} \in \mathbb{R}^{(h \cdot d_{k}) \times d}$.
Considering that aggregation occurs at every step, the sequence output from the MCA module is denoted by $x \in \mathbb{R}^{N \times d}$.
%
\subsubsection{Masked Multi-head Self Attention}
\label{subsubsec:masked_self_attention}
Originally introduced to address the neural machine translation problem, Transformer \cite{vaswani2017attention} has further improved the performance of deep learning approaches in several domains, and this method has been widely used in state-of-the-art studies.
The proposed concept of \emph{self-attention} maximizes the utilization of attention mechanisms \cite{bahdanau2014neural, luong2015effective}, capturing long-term correlations in given sequential data.
In WAY, the \emph{transformer decoder} architecture is adopted to convey past trajectory representations toward the current step and decode such information into features for destination estimation.

With the proceeded output $x \in \mathbb{R}^{N \times d}$ from the fore MCA module, MSA delivers the past aggregated information to the present.
The multi-headed attention weight $\alpha_{S} \in \mathbb{R}^{h \times N \times N}$ is defined by the following equations:
\begin{equation}
\label{eq:masked_multihead_self_attention}
\begin{aligned}
\begin{split}
Q_{i} &= xW^{Q}_{i} \text{\space\space} K_{i} = xW^{K}_{i} \text{\space\space} V_{i} = xW^{V}_{i} \\
\alpha_{S} &= \text{softmax}(\mathrm{F}_{mask}(\frac{QK^{T}}{\sqrt{d_{k}}})) \\
\Tilde{V} &= \alpha_{S} \cdot V 
\end{split}
\end{aligned}
\end{equation}
where $i \in [1, \dots ,h]$ indicates each head, $d_{k}$ denotes the head dimension, and $Q, K, \text{ and } V \in \mathbb{R}^{h \times N \times d_{k}}$ are the sets of vectors transformed by $W^{Q}_{i}, W^{K}_{i}, \text{ and } W^{V}_{i} \in \mathbb{R}^{d \times d_{k}}$, respectively.
Based on the matrix multiplication of $Q$ and $K^{T} \in \mathbb{R}^{h \times d_{k} \times N}$, the inter-correlation scores are measured over sequence steps, and the lower triangular matrix filter $\mathrm{F}_{mask}$ is applied to block information flows from future steps.
Thus, $\alpha_{S}$ is given, and $\Tilde{V} \in \mathbb{R}^{h \times N \times d_{k}}$ is produced, of which the vectors in $V$ are scaled and mixed with the representations from the past.
Eventually, the concatenation of the heads $\Tilde{V} \in \mathbb{R}^{N \times (h \cdot d_{k})}$ is multiplied by a linear transition $W^{out} \in \mathbb{R}^{(h \cdot d_{k}) \times d}$, providing the output sequence $\Tilde{V} \in \mathbb{R}^{N \times d}$.
%
\subsubsection{Channel Concatenation and Shared Feed Forward}
\label{subsubsec:shared_feed_forward}
Before applying SFF module procedure, the channel-wise concatenation between the original CASP layer input sequence $x \in \mathbb{R}^{C \times N \times d}$ and the output sequence $x \in \mathbb{R}^{N \times d}$ from MSA module proceeds.
The concatenation, which is more likely to be said to substitute a certain channel sequence, is to withdraw one of the channels from the original input and fill the spot with the output of MSA module.
Despite the replacement, the altered channel has more enriched information and incorporates the withdrawn identity because of the residual connections that will be described in Sec. \ref{sec:methodology}\ref{subsec:channel_aggregative_sequence_processing}\ref{subsubsec:residual_connection}.
In this study, the single channel is designated to be repeatedly replaced over the stack of CASP layers in the procedure, that is, the output from SE module.
The purpose of the design is to update a certain channel by gathering all channel-wise features and sequential information, whereas the others maintain their current step identities.
Based on the idea of CSPNet \cite{wang2020cspnet}, the connection not only relieves the computational burden, but also gives the model the ability to add up representations enriched repeatedly within the stack of CASP layers. 

Consequently, the concatenated tensor $x \in \mathbb{R}^{C \times N \times d}$ maintains the number of channels $C$, and SFF module performs linear transformations for each representation in a given tensor.
The module comprises 2-layered fully connected network $W_{1} \in \mathbb{R}^{d \times d_{f}}$ and $W_{2} \in \mathbb{R}^{d_{f} \times d}$ with intermediate ReLU activation, as the transition is shared in positions and channels.
Accordingly, CASP layer output $x \in \mathbb{R}^{C \times N \times d}$ can be obtained.
%
\subsubsection{Residual Connection}
\label{subsubsec:residual_connection}
As shown in Fig. \ref{fig:model_overview}, residual connection \cite{he2016deep} and layer normalization \cite{ba2016layer} follow each submodule in CASP layer.
The residual connection adds input $x$ to the output from the module, which is presented to address the vanishing gradient problem of deep neural network architectures.
Layer normalization is used to stabilize the hidden state dynamics in the sequential modeling.
As introduced in \cite{vaswani2017attention}, the same combination is applied to MSA and SFF, where the inputs and outputs of each have identical shapes.
However, in the MCA module, where channels are aggregated and the output comprises a single channel, only the designated input channel is subjected to such an architecture.
Considering that a certain channel was designated to be replaced in Sec. \ref{sec:methodology}\ref{subsec:channel_aggregative_sequence_processing}\ref{subsubsec:shared_feed_forward}, the channel to be withdrawn preserves its identity through residual connection, and moreover, it reflects the aggregated representation over both channels and sequence since MSA proceeds the output sequence from MCA module.
%
\subsection{Loss Function and Learning Technique}
\label{subsec:loss_function_and_learning_technique}
In this study, the classification problem of destination estimation is formulated given the vessel trajectory from its departure.
Therefore, let the model input be denoted as $x = \{x_{1}, \dots ,x_{T} ; Y_{x} ; S_{x} \}$, where an accumulated AIS observation from a single ongoing ship at time step $T$ and the voyage is departed from port $Y_{x}$ and the ship belongs to type $S_{x}$.
Thereafter, in processing the multi-channel sequence structure, WAY repeatedly aggregates information from channels and the past.
The current step of the designated channel representation $x \in \mathbb{R}^{d}$ is converted from the last CASP block output to a probability distribution over port classes using the final linear transition $W \in \mathbb{R}^{d \times Y}$ with the softmax function, which is defined as:
\begin{equation}
\label{eq:cross_entropy_loss}
\begin{aligned}
\begin{split}
\mathcal{L}_{CE}(y, x, \theta) &= -{ \log P(y | x_{1:T} ; Y_{x} ; S_{x}, \theta)}
\end{split}
\end{aligned}
\end{equation}
where $CE$ denotes Cross Entropy loss, $y \in Y$ is the future destination port, and $\theta$ denotes the model parameters.

\begin{figure}[!t]
\centering
\includegraphics[scale=0.55]{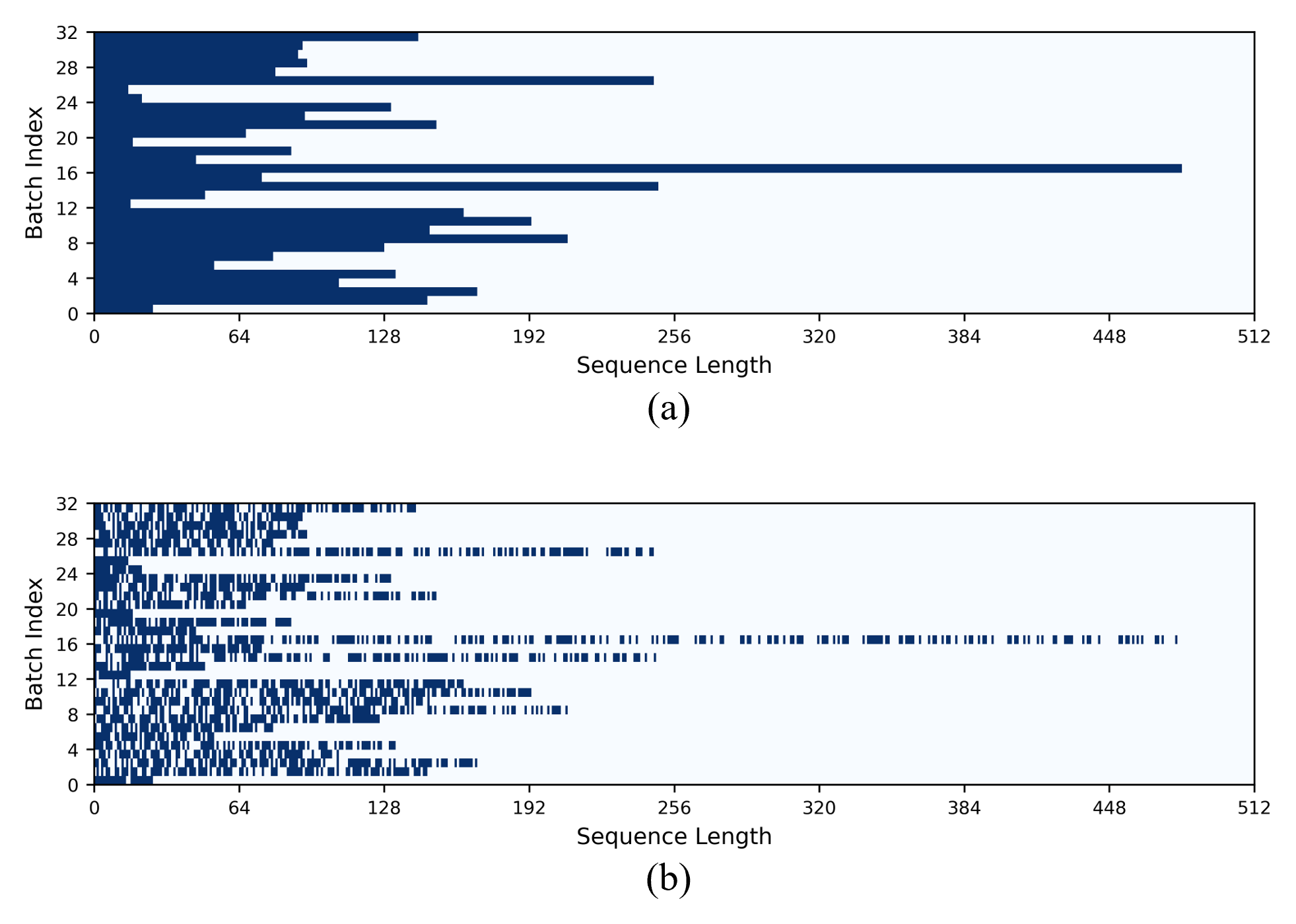}
\caption{An example of Gradient Dropout learning technique. (a) shows the valid steps of instances in the training batch and (b) visualizes the same batch after the technique is applied.}
\label{fig:gradient_dropout}
\end{figure}
Indeed, by exploiting the loss above, a many-to-many model training framework results in a biased parameter update.
Given the full port-to-port trajectories, the summation of feedback within the sequence toward a single destination can be affected by certain extremely long instances.
To address this issue, Gradient Dropout learning technique is proposed.
This technique controls the validity of stepwise loss from each instance in the mini-batch.
The validity is determined stochastically by the sampling ratio, which is inversely proportional to the log-scaled length of the instances, as follows:
\begin{equation}
\label{eq:gradient_dropout}
\begin{aligned}
\begin{split}
\delta_{k} &= 1 + \log_{\text{max}(N_{1}, \dots ,N_{B})}\frac{\text{min}(N_{1}, \dots ,N_{B})}{N_{k}}
\end{split}
\end{aligned}
\end{equation}
where $N_{k}$ for $k \in \{ 1, \dots , B \}$ denotes the length of the $k^{th}$ among $B$ instances in the mini-batch.
As shown in Fig. \ref{fig:gradient_dropout}, sampling ratio $\delta_{k}$ adjusts the feedback validity over instances, balancing the loss update against the lengths.
%
\section{EXPERIMENTS}
\label{sec:experiments}
%
\subsection{Dataset}
\label{subsec:dataset}
Satellite AIS data collected from ORBCOMM and port identification data provided by SeaVantage were used during this study.
Experiments were conducted on 5-year (January 2016-November 2020) accumulated real-world AIS data of 5,103 individual ships, where the vessels belong to one of three types: tanker, container, and bulk.
A total of $\simeq 130 K$ trajectories comprising $\simeq 17 M$ AIS messages were mapped between 3,243 port classes using the framework described in Section \ref{sec:problem_definition}\ref{subsec:annotation_framework}.
A destination-wise stratified split was used to generate the training/validation/test trajectory dataset as $70\%$, $15\%$, and $15\%$ of the total.

Given AIS trajectory instances, the features exploited by the proposed method are listed in Table \ref{tab:feature_table}.
In addition to the timestamps and coordinates, relevant to the preprocessing of the relative coordinates and time distances previously stated in Sec. \ref{sec:methodology}\ref{subsec:trajectory_representation}, other numerical features were adopted to reflect the local navigational patterns.
It should also be noted that ETA, which is human-generated, was preprocessed to the time distance between the current timestamp, and the departure port indicates the categorical index among the list of total port classes.
\begin{table}[h]
\caption{The components of AIS trajectory features.}
\label{tab:feature_table}
\centering
\begin{tabular}{L{1.5cm} L{3.8cm} L{1.5cm}}
\toprule
Category & Features & Unit \\
\midrule
\multirow{10}{*}{AIS} & Timestamp & UTC \\
                      & Estimated Time of Arrival & UTC \\
                      & Longitude & degree \\
                      & Latitude & degree \\
                      & Speed Over Ground & knot \\
                      & Rate Of Turn & degree \\
                      & Course Over Ground & degree \\
                      & Heading & degree \\
                      & AIS draught & meter \\
                      & Ship Type & categorical \\
\midrule
Port & Departure & categorical \\
\bottomrule
\end{tabular}
\end{table}
%
\subsection{Experimental Setup}
\label{subsec:experimental_setup}
%
\subsubsection{Benchmark Models and Implementation Details}
\label{subsubsec:benchmark_models_and_implementation_details}
Benchmark models selected from previous studies were employed to process AIS data, and the details were modified to meet the objective of the task.
LSTM \cite{hochreiter1997long} was the most frequently used sequence model in the maritime domain.
Sharing the gated sequential processing concept with LSTM, GRU \cite{cho2014learning} was occasionally adopted as a comparison in priors and the applications of the attention mechanism \cite{luong2015effective} to both LSTM and GRU were also considered.
In addition, considering that Transformer \cite{vaswani2017attention} has improved the performance of various sequential data processing fields, Transformer-decoder and TrAISformer \cite{nguyen2021traisformer} architectures were added for comparison.

Dividing the global ocean area into a spatial grid, a unit size of $1 \times 1$ (${}^{{\circ}^{2}}$) was set as the processing unit. 
As a contemplation of prior grid-token-based approaches, the comparison models embedded lookup tables for spatial vocabularies and learned their representations from the training set.
In the comparison of benchmarks, the models shared the hyperparameters of the number of stacked layers $L=4$ and hidden dimension size $d=128$.
In addition, TrAISformers, Transformer-decoder, and WAY shared the hyperparameters relevant to the attention architecture and feedforward network, which are the number of attention heads $h=4$, the head-wise hidden dimension size $d_{k}=64$, and the intermediate dimension size in the feedforward network $d_{f}=256$.
The models were trained using the Adam \cite{kingma2014adam} optimizer for $100$ epochs with a batch size of $32$ and dropout ratio of $30\%$, where the learning rate was set to $1e^{-4}$.
Test evaluations were conducted for each model when the minimal loss was recorded for the validation set.
%
\subsubsection{Evaluation Metric}
\label{subsubsec:evaluation_metric}
An accuracy metric was adopted to evaluate the models in terms of port classification, where the overall and quartile-based accuracies were measured.
The overall accuracy subjectifies the correctness of the model's estimation for every grid-unit step.
The evaluation of accuracy is measured as the equation below:
\begin{equation}
\label{eq:accuracy_metric}
\begin{aligned}
\begin{split}
\mathrm{Accuracy} &= \frac{\sum_{i=1}^{M}\sum_{j=1}^{N_{i}} (y_{i}=\hat{y}_{ij})}{\sum_{i=1}^{M}\sum_{j=1}^{N_{i}} \mathbbm{1}}
\end{split}
\end{aligned}
\end{equation}
where $i$ indexes the trajectory instances from the test set $M$, and $N_{i}$ corresponds to the length of unit-grid sequence composing the $i^{th}$ instance. The destination port class of $i^{th}$ trajectory is denoted by $y_{i}$, and $\hat{y}_{ij}$ indicates the class of the maximum probability the model infers at the progression step $j$ of the trajectory.
Given that the performance can vary depending on the voyage progress, the accuracy of different voyage progressions was also assessed.
Quartile-based accuracy was measured by adjusting the iteration range of $j$ from Equation \ref{eq:accuracy_metric}.
The models were evaluated along the quarter-wise accumulated progression of each test trajectory, hence the range of $(1, N_{i})$ at the term $\sum_{j=1}^{N_{i}}$ was adjusted into $(1, \frac{1}{4}N_{i})$, $(\frac{1}{4}N_{i}+1, \frac{1}{2}N_{i})$, $(\frac{1}{2}N_{i}+1, \frac{3}{4}N_{i})$, and $(\frac{3}{4}N_{i}+1, N_{i})$, according to quartiles respectively.

Additionally, an overall F1-score was adopted to verify the models under an imbalanced label condition. The measurement of F1-score for each port class $c$ follows the equation below:
\begin{equation}
\label{eq:f1_metric}
\begin{aligned}
\begin{split}
\mathrm{True\text{\space}Positive\text{\space\space}(TP)}&= 
\textstyle\sum_{i=1}^{M} \sum_{j=1}^{N_{i}} \mathbbm{1}^{c}_{y_{i}} (y_{i}=\hat{y}_{ij}) \\
\mathrm{False\text{\space}Positive\text{\space\space}(FP)} &= 
\textstyle\sum_{i=1}^{M} \sum_{j=1}^{N_{i}} \mathbbm{1}^{\slashed{c}}_{y_{i}} (c=\hat{y}_{ij}) \\
\mathrm{False\text{\space}Negative\text{\space}(FN)}&= \textstyle\sum_{i=1}^{M} \sum_{j=1}^{N_{i}} \mathbbm{1}^{c}_{y_{i}} (y_{i} \neq \hat{y}_{ij}) \\
\text{\space\space\space\space\space\space\space\space\space\space\space\space\space\space}
\mathrm{F1\text{-}score}&=\frac{\mathrm{TP}}{\mathrm{TP} + 0.5(\mathrm{FP} + \mathrm{FN})}
\end{split}
\end{aligned}
\end{equation}
where $\mathbbm{1}^{c}_{y_{i}}$ denotes if destination port label $y_{i}$ is class $c$ and $\mathbbm{1}^{\slashed{c}}_{y_{i}}$ denotes that the destination of the $i^{th}$ trajectory is other than the class $c$. Finally, the overall measure of F1-score for the multi-class classification problem is computed by averaging the class-wise F1-scores, also called macro-averaged F1-score.
\begin{table*}[!t]
\caption{Benchmark comparison results. The models are evaluated by quartered progression (Q) and overall trajectory steps. Quartile division is based on the length of the observed data.}
\label{tab:classification_results}
\centering
\begin{tabular}{L{3.8cm} C{1.8cm} C{1.6cm} C{1.6cm} C{1.6cm} C{1.6cm} c C{1.8cm}}
\toprule
\multirow{2.5}{*}{Method} & \multicolumn{5}{c}{Accuracy(\%)}                                  & & F1-score(\%) \\ 
                            \cmidrule{2-6}                                                        \cmidrule{8-8}
                          & {\bf Overall} 
                          & 0Q $\backsim$ 1Q 
                          & 1Q $\backsim$ 2Q 
                          & 2Q $\backsim$ 3Q 
                          & 3Q $\backsim$ 4Q
                          & & {\bf Overall} \\
\midrule
LSTM                        & 56.48 $\pm$0.32 & 40.90 $\pm$0.60 & 56.63 $\pm$0.16 & 62.48 $\pm$0.27 & 67.04 $\pm$0.49 & & 27.30 $\pm$1.22 \\
LSTM-ATTN                   & 55.76 $\pm$0.17 & 41.60 $\pm$0.15 & 55.77 $\pm$0.35 & 61.56 $\pm$0.35 & 65.13 $\pm$0.23 & & 27.40 $\pm$0.82 \\
GRU                         & 56.35 $\pm$0.40 & 41.30 $\pm$0.07 & 55.74 $\pm$0.55 & 62.18 $\pm$0.50 & 67.25 $\pm$0.52 & & 29.02 $\pm$0.11 \\
GRU-ATTN                    & 56.45 $\pm$0.09 & 41.54 $\pm$0.26 & 56.32 $\pm$0.39 & 62.25 $\pm$0.16 & 66.60 $\pm$0.21 & & 28.21 $\pm$0.15 \\
Transformer-decoder         & 60.26 $\pm$0.05 & 44.35 $\pm$0.26 & 59.45 $\pm$0.27 & 65.83 $\pm$0.32 & 72.46 $\pm$0.12 & & 32.79 $\pm$0.53 \\
TrAISformer                 & 64.38 $\pm$0.14 & 50.15 $\pm$0.22 & 63.62 $\pm$0.10 & 69.44 $\pm$0.12 & 75.27 $\pm$0.13 & & 33.12 $\pm$0.26 \\
TrAISformer-multiResolution & 64.60 $\pm$0.28 & 50.23 $\pm$0.36 & 64.07 $\pm$0.49 & 69.60 $\pm$0.22 & 75.46 $\pm$0.13 & & 34.10 $\pm$0.26 \\
\midrule
WAY (ours)                  & {\bf 79.45} $\pm$0.32 & 71.21 $\pm$0.21 & 79.50 $\pm$0.22 & 82.42 $\pm$0.40 & 85.25 $\pm$0.47 & & {\bf 49.45} $\pm$0.68 \\
WAY w/ GD (ours)            & {\bf 80.44} $\pm$0.11 & 71.96 $\pm$0.20 & 80.19 $\pm$0.14 & 83.38 $\pm$0.06 & 86.81 $\pm$0.20 & & {\bf 52.01} $\pm$0.80 \\
\bottomrule
\end{tabular}
\end{table*}
\begin{figure*}[!t]
\centering
\includegraphics[scale=0.7]{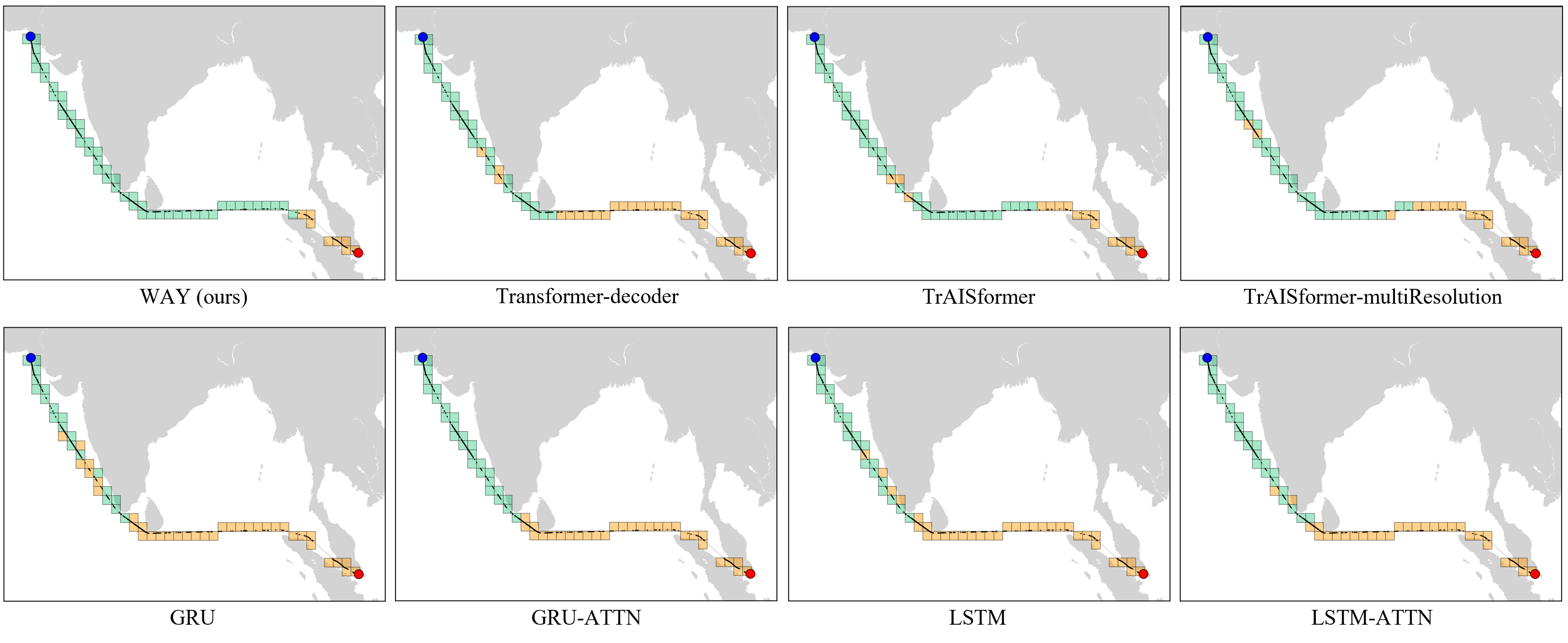}
\caption{The actual estimation result of each comparison model for the sample test trajectory. The trajectory proceeds from the \textcolor{red}{Red} marker (Departure port) to the \textcolor{blue}{Blue} marker (Destination port). The grids are colored \textcolor[rgb]{0, 0.7, 0.4}{Green} if the estimation was correct at the respective steps, while the \textcolor[rgb]{1, 0.55, 0}{Orange} denotes the model has estimated the wrong destination.}
\label{fig:model_estimation_example}
\end{figure*}
%
\subsection{Experimental Results}
\label{subsec:experimental results}
\begin{table*}[!t]
\caption{Verifying the effect of Gradient Dropout (GD) learning technique.}
\label{tab:gradientdropout_comparison}
\centering
\begin{tabular}{L{4cm} C{1.7cm} C{1.7cm} C{0.5cm} C{1.7cm} C{1.7cm}}
\toprule
\multirow{2.5}{*}{Method} &  \multicolumn{2}{c}{Accuracy(\%)} & & \multicolumn{2}{c}{F1-score(\%)} \\ 
                             \cmidrule{2-3}                       \cmidrule{5-6}
                          & w/o & w/ GD                       & & w/o & w/ GD \\
\midrule
LSTM                        & 56.48 $\pm$0.32 & 57.20 $\pm$0.30 & & 27.30 $\pm$1.22 & 28.85 $\pm$0.47 \\
LSTM-ATTN                   & 55.76 $\pm$0.17 & 56.48 $\pm$0.10 & & 27.40 $\pm$0.82 & 29.22 $\pm$0.20 \\
GRU                         & 56.35 $\pm$0.40 & 57.70 $\pm$0.25 & & 29.02 $\pm$0.11 & 30.23 $\pm$0.77 \\
GRU-ATNN                    & 56.45 $\pm$0.09 & 57.15 $\pm$0.16 & & 28.21 $\pm$0.15 & 29.96 $\pm$0.28 \\
Transformer-decoder         & 60.26 $\pm$0.05 & 60.68 $\pm$0.07 & & 32.79 $\pm$0.53 & 33.22 $\pm$0.10\\
TrAISformer                 & 64.38 $\pm$0.14 & 66.06 $\pm$0.46 & & 33.12 $\pm$0.26 & 36.30 $\pm$0.35\\
TrAISformer-multiResolution & 64.60 $\pm$0.28 & 66.13 $\pm$0.03 & & 34.10 $\pm$0.26 & 37.00 $\pm$0.06\\
\midrule
WAY (ours) & 
{\bf 79.45} $\pm$0.32 & {\bf 80.44} $\pm$0.11 & & {\bf 49.45} $\pm$0.68  & {\bf 52.01} $\pm$0.80 \\
\bottomrule
\end{tabular}
\end{table*}
Two types of experimental results are presented in this subsection: First, the overall performance of estimating the long-term destination between WAY and other benchmark models is compared, whilst another experiment verifies the advantage of using Gradient Dropout (GD).

As shown in Table \ref{tab:classification_results}, WAY achieves a remarkable improvement compared to the other benchmark models, and WAY with GD further demonstrates a performance gain.
As well as the overall scores, WAY improves the performance of destination estimation for all steps along the trajectory, especially of which steps were included before the $1^{st}$ quartile progression. Fig. \ref{fig:model_estimation_example} illustrates the example of actual estimation results from each comparison model given the sample trajectory.
In this regard, the experiment implies the limitation of the coarse spatial representation from the basic grid-token approaches for handling AIS data worldwide.
This suggests the importance of enriching the detailed representations of the AIS trajectory and aggregating information effectively.

Among the conventional methods that represent trajectories using only spatial tokens (LSTM, GRU, -ATTNs, Transformer-decoder), Transformer–decoder records the highest performance.
Differentiated from the other prior recurrent sequence processing models, which are considerably dependent on the current step representation, Transformer–decoder is shown to be advantageous for handling a long sequence of AIS trajectories.
\emph{Self attention}, the concept introduced in Transformer to maximize the utility of the attention mechanism, is robust to the long-term dependency while treating each step equally in the processing of sequential steps.
Contrastively, the appliance of the attention mechanism, on the top of the recurrent processing models, is shown to be uncompetitive. Considering that the objective is to estimate long-range destinations far beyond time and space, the recurrence is speculated as rather enforcing the dependency and failing to capture the inter-token long-term correlation along the trajectory. Thus, it seems difficult that the improvement is anticipated using a shallow attention computation under lack of sharable representation.

TrAISformers, which have an identical structure to Transformer-decoder, accomplish extra improvement by carrying out their own tokenization method that accepts extra features from AIS trajectories other than the prevalent spatial tokenization.
In the approach of TrAISformers, the regions are tokenized separately by each axis of longitude and latitude, and SOG and COG are discretized and also represented as tokens.
Thereafter, they concatenate the representations of the coordinates, SOG, and COG, allowing the \emph{transformer decoder} to process the data.
The improvement shown in TrAISformers supports that the availability of AIS features other than spatial coordinates, enriching the representations of AIS trajectory, benefits the performance despite adopting the same sequential processing steps.

The results also show discrepancies between the accuracy and F1 measure from the benchmark models, including WAY.
This is attributed to the disparities originating from the imbalance in port-to-port operation frequency depending on the ship type.
Nevertheless, this study confirms that WAY clearly manifests its performance on both sides of the evaluation.

The results in Table \ref{tab:classification_results} verify that GD increases the performance of WAY.
To generalize the advantage of the task-specialized learning technique in the trajectories with large length deviations, GDs were applied to benchmark models.
As shown in Table \ref{tab:gradientdropout_comparison}, the results suggest that applying GD always achieves a performance gain regardless of the backbone model.
The average gains are $+1.01\%$ for the overall accuracy and $+1.92\%$ for the F1-score.
%
\subsection{Further Experiments}
\label{subsec:further_experiments}
In this subsection, further experiments are conducted to examine the effects of the proposed methods on the performance.
The experiments covers the comparison results based on the feature types, channel aggregation methods, and model capacity.

\noindent \textbf{Feature types.}
The importance of considering trajectory representations are discussed, where, to achieve enriched representations, WAY divided the feature types, other than spatial areas, into local navigational patterns, departure, and ship types and developed representation methods to fit each type.
In Table \ref{tab:channel_comparison}, an experiment is conducted to investigate the effect of each feature type by removing one feature type.
It is observed that all of the feature types significantly affected the performance gain, and navigational patterns, particularly contributed to the improvement in the performance.

\noindent \textbf{Channel aggregation methods.}
Considering the multi-channel representations of AIS trajectories, aggregation methods for inter-channel information are explored.
Unifying channels via dimension-wise concatenation is a typical method for aggregating information distributed in multiple channels.
Introduced to process multi-modal time series, Cross Attention \cite{tsai2019multimodal} is a representative method to reflect mutual information between sequences.
In addition, the original Channel Attention \cite{hu2018squeeze}, the predecessor of MCA module in WAY is also examined without the extension of multi-headed perspectives.
The results listed in Table \ref{tab:aggregation_comparison} suggest that the proposed channel aggregation method is outstanding in terms of assembling the representations of each channel.
\begin{table}[!t]
\caption{Comparison of performance dependence on each represented channel.}
\label{tab:channel_comparison}
\centering
\begin{tabular}{L{3.5cm} C{1.75cm} C{1.75cm}}
\toprule
Channel Availability & Accuracy(\%) & F1-score(\%) \\
\midrule
w/o Local Pattern    & 64.70 $\pm$0.24 & 38.31 $\pm$0.57 \\
w/o Departure  & 76.35 $\pm$0.46 & 42.36 $\pm$0.64 \\
w/o Ship Type  & 78.84 $\pm$0.22 & 48.58 $\pm$0.78 \\
\midrule
Use All (base) & {\bf 79.45} $\pm$0.32 & {\bf 49.45} $\pm$0.68 \\
\bottomrule
\end{tabular}
\end{table}
\begin{table}[!t]
\caption{Comparison of inter-channel aggregation methods.}
\label{tab:aggregation_comparison}
\centering
\begin{tabular}{L{3.5cm} C{1.75cm} C{1.75cm}}
\toprule
Aggregation Method & Accuracy(\%) & F1-score(\%) \\
\midrule
Concatenation                & 73.59 $\pm$0.10 & 42.47 $\pm$0.27 \\
Cross Attention              & 75.61 $\pm$0.49 & 40.95 $\pm$0.10 \\
Channel Attention            & 77.94 $\pm$0.34 & 48.59 $\pm$0.49 \\
\midrule
Multi-head Channel Attention & {\bf 79.45} $\pm$0.32 & {\bf 49.45} $\pm$0.68 \\
\bottomrule
\end{tabular}
\end{table}
\begin{table}[!t]
\caption{Comparison of model capacity and the performances.}
\label{tab:modelsize_comparison}
\centering
\begin{tabular}{L{1.55cm} C{1.55cm} C{1.75cm} C{1.75cm}}
\toprule
Method & \#Parameter & Accuracy(\%) & F1-score(\%) \\
\midrule
LSTM        & 4.76M & 56.48 $\pm$0.32 & 27.30 $\pm$1.22 \\
GRU         & 4.62M & 56.35 $\pm$0.40 & 29.02 $\pm$0.11 \\
TF-decoder  & 5.02M & 60.26 $\pm$0.05 & 32.79 $\pm$0.53 \\
TrAISformer & 1.30M & 64.38 $\pm$0.14 & 33.12 $\pm$0.26 \\
\midrule
WAY-Tiny    & {\bf 0.24M} & 69.51 $\pm$0.22 & 36.32 $\pm$0.28 \\
WAY-Small   & 0.52M       & 74.68 $\pm$0.05 & 43.92 $\pm$0.66 \\
WAY (base)  & 2.04M       & {\bf 79.45} $\pm$0.32 & {\bf 49.45} $\pm$0.68 \\
\bottomrule
\end{tabular}
\end{table}

\noindent \textbf{Model capacity.}
Finally, Table \ref{tab:modelsize_comparison} presents the experimental results of the model comparison in terms of capacity and performance.
The experiment includes benchmark models with a conventional grid-token-based approach (LSTM, GRU, and Transformer-decoder), TrAISformer, which designed its own technique of token-based representation, and WAY-Tiny and WAY-Small.
WAY-Tiny and WAY-Small have the same architecture as WAY (base); however, their hyperparameters are adjusted, where $\{ L, d, h, d_{k}, d_{f} \}$ are set to $\{ 2, 32, 2, 16, 128 \}$ and $\{ 2, 64, 2, 32, 128 \}$, respectively.
The results demonstrates that the previous methods for laying an embedding table of learnable parameters for spatial areas overwhelm the number of parameters, while their performances are uncompetitive.
TrAISformer succeeds in reducing parameters and improving the performance compared to the above but fails to overcome the parameterizing for spatial correlation.
However, WAY (base) achieves the highest score evaluation without wasting a massive number of parameters for covering worldwide regions.
Although WAY-Tiny and WAY-Small are not shown to have the best performance, they still outperform the other benchmarks with $2.5 \sim 20.5$ times fewer parameters.
%
\section{DISCUSSION}
\label{sec:discussion}
This section discusses additional tasks before applying WAY to solve the port congestion problem.
As proved in Section \ref{sec:experiments}, WAY achieved outstanding performance in terms of destination estimation; however, knowing both the destination and arrival times of vessels is necessary to address the port congestion problem.
Considering its application in the industry, a further experiment is conducted to verify whether WAY, as an end-to-end model, could show outstanding performance in terms of both destination and arrival time estimation.
WAY-Mul, comprising the classifier and an additional regression head to estimate both the destination and arrival time, is designed and trained with the same parameters as used for WAY. 

Table \ref{tab:eta_experiment} summarizes the comparison result between human-made ETA and WAY-Mul's estimation.
WAY-Mul reduces the arrival time errors from $4.26$ days to $2.90 \sim 3.03$ days while keeping the performance at destination estimation.
Although the experiment result shows WAY's expandability from single-task to multi-task learning, the performance of estimating arrival time needs to be improved to solve the port congestion problem.
It is shown that a low performance of arrival time estimation occurred owing to inconsistent arrival time labels.
The arrival time of vessels is affected by voyage conditions such as the distance between departure and destination port, vessel speed, and draft.
However, external factors like port congestion and due date of shipping result in waiting time, which leads to inconsistent arrival time.
Since WAY's expandability to multi-task learning is verified, WAY can achieve outstanding performance in terms of both destination and arrival time estimation if consistent arrival time labels are guaranteed.
In future studies, a different annotation framework needs to be designed for consistent arrival time labels.
Given the consistent arrival time labels, the authors believe that the accurate estimation within both destination and arrival time can be achieved, and thus this study contributes to mitigating the port congestion problem.
\begin{table}[!t]
\caption{The comparison results in terms of destination and arrival time estimation.}
\label{tab:eta_experiment}
\centering
\begin{tabular}{lcccc}
\toprule
\multirow{2.5}{*}{Method} & \multicolumn{2}{c}{Objective} & \multicolumn{2}{c}{Metric} \\
\cmidrule(r){2-3} \cmidrule(l){4-5} & Destination & Arrival time & Accruracy & MAE \\ 
\midrule
Human-made &             & \checkmark & -     & 4.26 \\
WAY        &  \checkmark &            & 79.45 & -  \\
WAY GD     &  \checkmark &            & 80.44 & -  \\
\midrule
WAY-Mul    & \checkmark  & \checkmark & 79.30 & 3.03 \\
WAY-Mul GD & \checkmark  & \checkmark & 80.05 & 2.90 \\
\bottomrule
\end{tabular}
\end{table}
%
\section{CONCLUSION}
\label{sec:conclusion}
This study establishes the necessity of handling global range AIS data to address the problem of port destination estimation and introduces a novel data-driven approach toward the AIS trajectory. 
Based on the premise that a port-to-port annotated trajectory is given, the objective of maximizing the probability of the target destination is formulated. 
Recasting the objective defined on message-wise trajectory processing to the uniform grid-wise processing, the approach mitigates the spatio-temporal bias within a long international AIS data forming irregular intervals. Concurrently, by adopting the nested sequence representation, the rearranged trajectory preserves the details of the original.

Based on this approach, the end-to-end model architecture of WAY is proposed, which demonstrates state-of-the-art performance by processing the AIS trajectory as a 4-channel vector sequence. Based on a multi-channel representation, WAY processes AIS trajectories in spatial grid steps without losing the detailed AIS properties such as global-spatial identities, local patterns, semantics, and time irregular progression of trajectory. Experimental results highlight the importance of enriching the details and effective aggregation of the long sequence while showing the limitations of the conventional grid-based approach, that is, the coarse portrayal of trajectories.

Additionally, the experimental results verify that the adoption of Gradient Dropout, a task-specialized learning technique, leads to further performance gain in the estimation.
The performance gain is shown not only for WAY, but also for other benchmarks based on the conventional spatial grid approach.

Finally, this study shows the expandability of WAY into multi-tasked learning for ETA, and discusses a provision of real-world port congestion estimation that can be realized through further improvement in identifying the accurate time of arrival from historical AIS data.
%
\bibliographystyle{IEEEtran}
\bibliography{References}
%
\begin{IEEEbiography}[{\includegraphics[width=1in,height=1.25in,clip,keepaspectratio]{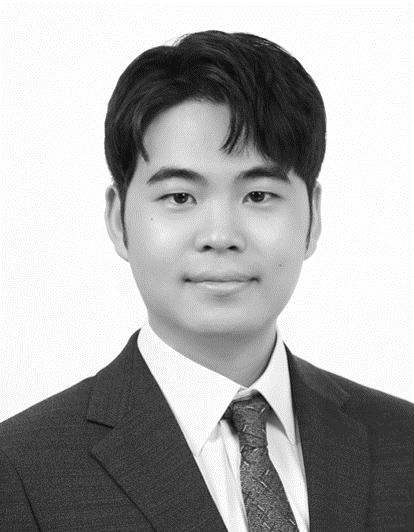}}]{Jin Sob Kim}
{\space}received the B.S. degree in computer engineering from Hongik University, Seoul, Republic of Korea, in 2021. He is currently pursuing a Ph.D. degree at the school of industrial and management engineering with Korea University, Seoul, Republic of Korea. 

His research interests include spatio-temporal data mining, attention mechanism, time series modeling, and deep learning.
\end{IEEEbiography}
\begin{IEEEbiography}[{\includegraphics[width=1in,height=1.25in,clip,keepaspectratio]{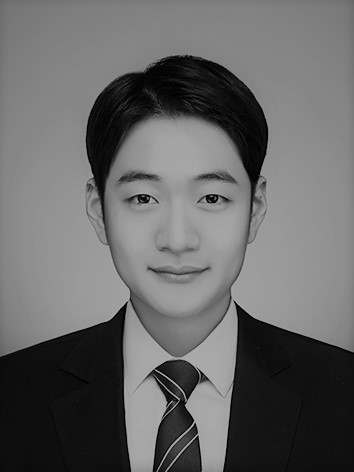}}]{Hyun Joon Park}
{\space}received the B.S. degree in industrial engineering from Hongik University, Seoul, Republic of Korea, in 2020. He is currently pursuing a Ph.D. degree at the school of industrial and management engineering at Korea University, Seoul, Republic of Korea. 

His research interests include attention mechanism, time series modeling, speech enhancement, and deep learning.
\end{IEEEbiography}
\begin{IEEEbiography}[{\includegraphics[width=1in,height=1.25in,clip,keepaspectratio]{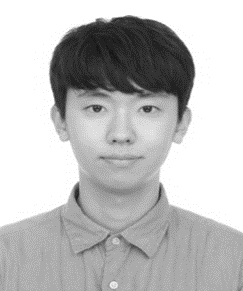}}]{Wooseok Shin}
{\space}received the B.S. degree in industrial and information systems engineering from Seoul National University of Science and Technology, Seoul, Republic of Korea, in 2020. He is currently pursuing a Ph.D. degree at the school of industrial and management engineering with Korea University, Seoul, Republic of Korea. 

His research interests include model compression in computer vision and speech.
\end{IEEEbiography}
\begin{IEEEbiography}[{\includegraphics[width=1in,height=1.25in,clip,keepaspectratio]{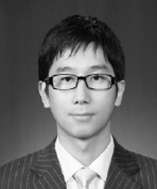}}]{Dongil Park}
{\space}received the B.S. degree in computer engineering from Gyeongnam National University of Science, Gyeongnam, Korea, in 2003. He is currently in charge of Chief Technical Officer at SeaVantage, in Seoul, Korea. 

His research interests include the development of maritime platforms, ship routing systems, transportation economics, and applications of deep learning in the vessel domain.
\end{IEEEbiography}
\begin{IEEEbiography}[{\includegraphics[width=1in,height=1.25in,clip,keepaspectratio]{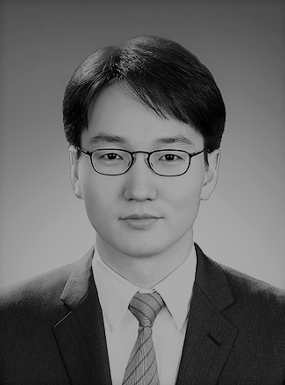}}]{Sung Won Han}
{\space}received the M.S. degrees in operations research, in statistics, and in mathematics from the Georgia Institute of Technology, Atlanta, GA, USA in 2006, 2007, and 2010, respectively, and the Ph.D. degree from the school of industrial and systems engineering (statistics), Georgia Institute of Technology, Atlanta, GA, USA. He was a senior research scientist with the division of biostatistics, school of medicine, New York University, NY, USA, and a postdoctoral researcher with the department of biostatistics and epidemiology/center for clinical epidemiology and biostatistics, school of medicine, University of Pennsylvania, Philadelphia, PA, USA. He is currently an associate professor with the school of industrial and management engineering, Korea University, Seoul, Republic of Korea. 

His research interests include probabilistic graphical model, network analysis, machine learning, and deep learning.
\end{IEEEbiography}

\vfill
\pagebreak
\clearpage

\appendix
%
\subsection{Port Label Annotation}
\label{appendix:port_label_annotation}
\begin{algorithm}[!h]
\caption{Destination candidate extraction}
\label{alg:destination_candidate_extraction}
\begin{algorithmic}
\STATE {\textsc{Regularize}}$(exp[1:m])$
\STATE \hspace{0.5cm} $ \textbf{for } i \gets 1 \text{ to } m \textbf{ do } $
\STATE \hspace{1.0cm} $ exp[i] \gets \text{space} \textbf{ if } exp[i] \in \text{special character} $
\STATE \hspace{1.0cm} $ \text{strip}(exp[i]) \textbf{ if } exp[i-1] = \text{space} $
\STATE \hspace{0.5cm} $ exp \gets \text{N-gram}(exp, \text{space}) $
\STATE \hspace{0.5cm} $ \textbf{return } exp $
\STATE
\STATE {\textsc{Extract Candidate}}$(exp\{1:N\}, port\{1:Y\})$
\STATE \hspace{0.5cm} $ candidates \gets \emptyset $
\STATE \hspace{0.5cm} $ \textbf{for } n \gets 1 \text{ to } N \textbf{ do } $
\STATE \hspace{1.0cm} $ \textbf{for } y \gets 1 \text{ to } Y \textbf{ do } $
\STATE \hspace{1.5cm} $ distance \gets DL(exp[n], port[y]) $
\STATE \hspace{1.5cm} $ score \gets 1 - (distance / |exp[n]|) $
\STATE \hspace{1.5cm} $ \textbf{if } score > thresh \textbf{ then} $
\STATE \hspace{2.0cm} $ candidates \gets candidates \cup \{ port[y] \} $
\STATE \hspace{0.5cm} $ \textbf{return } candidates $
\end{algorithmic}
\label{alg1}
\end{algorithm}
The AIS destination, string data written by humans, semantically indicates the destination port of the current navigation.
However, the data have been barely trustworthy because of the errors caused by humans paradoxically, such as unformatted indications, diverse languages, misspells, delaying updates, and unrelated information to destinations.
In many cases, however, the data have alphabetical similarities with port identifications, which are the names and UN/LOCODE (code for ports and other locations).
Therefore, the resemblances support the idea that the data contain an intention to express the destination in its own way while the freight is ongoing.
To read the semantic context of the written data and map it to the actual ports, Damerau--Levenshtein (DL) distance \cite{levenshtein1966binary, damerau1964technique} method is used to measure the similarity between the AIS destination and port identifications. 
The thresholding of similarities yields the current destination candidates. 
The posterior verification is then conducted on nominee ports based on whether future trajectory progression illustrates that the ship actually arrives and docks to the ports.

For each transmitted AIS message, given AIS destination expression $exp$ comprising $m$ characters, the Algorithm \ref{alg:destination_candidate_extraction} is the pseudocode for computing the similarities and narrowing down the destination candidates among $Y$ port identifications.
The expression is first regularized and turned into an N-gram set to consider any possibility of partial context being included.
Thereafter, for individual grams in the set, the candidates are extracted by similarity scores computed for each port identification, where the scores are measured based on the DL distance, $|exp[n]|$ denotes the length of the $n^{th}$ gram string, and $thresh=0.75$.
\begin{algorithm}[!t]
\caption{Posterior candidate validation}
\label{alg:posterior_candidate_validation}
\begin{algorithmic}
\STATE {\textsc{Positional Status}}$(\mathbf{X}\{1:T\}, port\{1:Y\})$
\STATE \hspace{0.5cm} $ status \gets \emptyset $
\STATE \hspace{0.5cm} $ \textbf{for } t \gets 1 \text{ to } T \textbf{ do } $
\STATE \hspace{1.0cm} $ coord \gets \mathbf{X}[t].(Longitude, Latitude) $
\STATE \hspace{1.0cm} $ \textbf{if } status[-1] = c \textbf{ then} $
\STATE \hspace{1.5cm} $ distance \gets Euclidean(coord, center) $
\STATE \hspace{1.5cm} $ \textbf{if } rad > distance \textbf{ then} $
\STATE \hspace{2.0cm} $ \text{append}(status, c); \text{ } \textbf{continue} $
\STATE \hspace{1.0cm} $ flag \gets \text{false} $
\STATE \hspace{1.0cm} $ \textbf{for } c \gets \mathbf{X}[t].candidates\{1:C\} \textbf{ do } $
\STATE \hspace{1.5cm} $ rad \gets \alpha \sqrt{\text{area}(port[c].Polygon) / \pi} $
\STATE \hspace{1.5cm} $ center \gets \text{center}(port[c].Polygon) $
\STATE \hspace{1.5cm} $ distance \gets Euclidean(coord, center) $
\STATE \hspace{1.5cm} $ \textbf{if } rad > distance \textbf{ then} $
\STATE \hspace{2.0cm} $ flag \gets \text{true} $
\STATE \hspace{2.0cm} $ \text{append}(status, c); \text{ } \textbf{break} $
\STATE \hspace{1.0cm} $ \textbf{if } flag = \text{true} \textbf{ then} $
\STATE \hspace{1.5cm} $ \textbf{continue} $
\STATE \hspace{1.0cm} $ \textbf{if } \mathbf{X}[t].SpeedOverGround > 1 \textbf{ then} $
\STATE \hspace{1.5cm} $ \text{append}(status, 0) $
\STATE \hspace{1.0cm} $ \textbf{else } $
\STATE \hspace{1.5cm} $ \text{append}(status, \text{null}) $
\STATE \hspace{0.5cm} $ \textbf{return } status $
\STATE
\STATE {\textsc{Extract Trajectory}}$(\mathbf{X}\{1:T\})$
\STATE \hspace{0.5cm} $ \mathbf{S} \gets \emptyset; \text{ } s \gets \emptyset $
\STATE \hspace{0.5cm} $ \textbf{for } t \gets 1 \text{ to } T \textbf{ do } $
\STATE \hspace{1.0cm} $ \textbf{if } \mathbf{X}[t].status \in \{ 0, \text{null} \} \textbf{ then} $
\STATE \hspace{1.5cm} $ \text{append}(s, \mathbf{X}[t]); \text{ } \textbf{continue} $
\STATE \hspace{1.0cm} $ j \gets 0 $
\STATE \hspace{1.0cm} $ port \gets \mathbf{X}[t].status \in port\{ 1:Y \} $
\STATE \hspace{1.0cm} $ \textbf{while } \mathbf{X}[t+j].status = port \textbf{ do } $
\STATE \hspace{1.35cm} $ \text{append}(s, \mathbf{X}[t+j]) $
\STATE \hspace{1.35cm} $ \textbf{if } \text{nearest}(\mathbf{X}[t+j].coord, port.center) \textbf{ then}$
\STATE \hspace{2.0cm} $ \text{append}(\mathbf{S}, s); \text{ } s \gets \emptyset $
\STATE \hspace{1.35cm} $ j \gets j+1 $
\STATE \hspace{1.0cm} $ \textbf{continue as } t \gets t+j $
\STATE \hspace{0.5cm} $ \textbf{return } \mathbf{S} $
\STATE
\STATE {\textsc{Validate Trajectory}}$(\mathbf{S}\{1:U\})$
\STATE \hspace{0.5cm} $ \textbf{for } s\{ t \gets 1:\hat{T} \} \gets S\{ 1:U \} \textbf{ do } $
\STATE \hspace{1.0cm} $ dept \gets s[1].status; \text{ } dest \gets s[\hat{T}].status $
\STATE \hspace{1.0cm} $ \textbf{if } dept, dest \notin \{ 0, \text{null} \} $
\STATE \hspace{1.0cm} $ \textbf{ and } dept \neq dest $
\STATE \hspace{1.0cm} $ \textbf{ and } \forall{s[t].status} \neq \text{null} $
\STATE \hspace{1.0cm} $ \textbf{ and } \exists{s[t].status} = 0 $
\STATE \hspace{1.0cm} $ \textbf{ and } \forall{s[t].candidates} \cap \{ dept, dest \} \neq \emptyset $
\STATE \hspace{1.0cm} $ \textbf{ and } \forall{(s[t] - s[t-1]).Timestamp} < thresh $
\STATE \hspace{1.5cm} $ \textbf{continue} $
\STATE \hspace{1.0cm} $ \text{delete}(\mathbf{S}, s) $
\STATE \hspace{0.5cm} $ \textbf{return } \mathbf{S} $
\end{algorithmic}
\label{alg2}
\end{algorithm}

Subsequently, based on the candidate nominated ports, valid trajectory continuities are verified by kinematic features at the posterior perspective regarding any range of sequences from raw AIS data.
The pseudocode Algorithm \ref{alg:posterior_candidate_validation}, comprising three subprocedures, describes the procedure of annotating trajectories from the raw accumulated data $\mathbf{X}$ and the verification of each annotated trajectory such that the set of valid trajectories $\mathbf{S}$ remains at the end.

Before the start of this procedure, the authors strongly recommend the readers to use spherical distance measure (e.g., Haversine) replacing the $Euclidean$ in Algorithm \ref{alg:posterior_candidate_validation}.
Since it rather fits to assume the coordinates exist on a spherical surface than in two-dimensional space.

\begin{figure}[!t]
\centering
\includegraphics[scale=0.65]{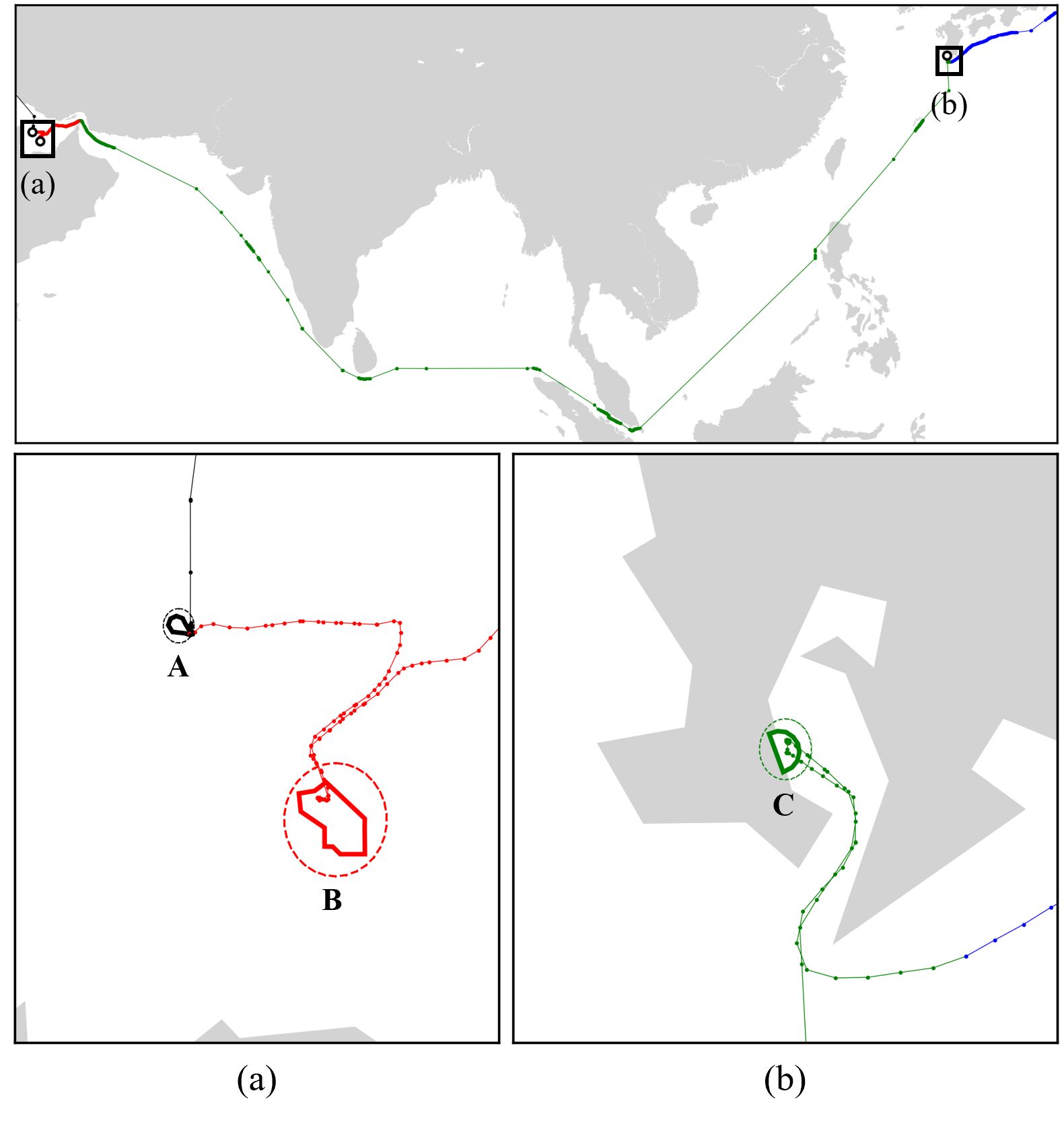}
\caption{Visualization of posterior candidate validation in-process. The voyage proceeds from left to right, and colors denote observations with the port as one of the candidates. (a) and (b) are expanded images from the box regions above, illustrating the geo-polygons of each port and their thresholds around these areas.}
\label{fig:posterior_candidate_validation}
\end{figure}

The first part verifies the candidates extracted from Algorithm \ref{alg:destination_candidate_extraction} and confirms the actual moment of arrival within the coordinates and SOG recorded in the raw data.
This process generates status tags for positions on whether the ship is on the dock or ongoing.
The border of tag decisions is dependent on the size of the geo-polygon areas marginalized by $\alpha=1.8$, which creates circular boundaries, such as the dashed lines around each port area shown in Fig. \ref{fig:posterior_candidate_validation}a and Fig. \ref{fig:posterior_candidate_validation}b.
In addition, under the circumstances of an ongoing ship, two status tags are specified, where 0 indicates acceleration and the null value denotes staying still at the location.
These tags are used to validate the trajectories.

Referring to the positional status tags defined in the previous process, the accumulated data are split into individual sections whose departure and destination were confirmed.
The division occurs at the nearest point among the observations included in the port boundaries, organizing the set of trajectories $\mathbf{S}$.
Finally, the trajectories are verified with a singular pair of departure and destination.
In other words, the sections with the possibility of the subject containing more than one ship operation are excluded; however, the pairs at the beginning and the end were confirmed.
In this validation, the aforementioned status tags are used to affirm whether a trajectory sequence included a cease behavior apart from a confirmed departure or destination location, while the acceleration must have been shown in the trajectory at least once.
In addition, the context of the human source is verified over the sequences, checking of port candidates from every observation if any moment exists that neither candidate intended to represent departure or destination.
The insufficiency of the kinetical reason forms part of the validation, and $thresh=3 \textit{ days}$ is set as the maximum allowance of the time gap between continuous observations.
%
\subsection{Trajectory Refinement}
\label{appendix:trajectory_refinement}
\begin{figure}[!t]
\centering
\includegraphics[scale=0.60]{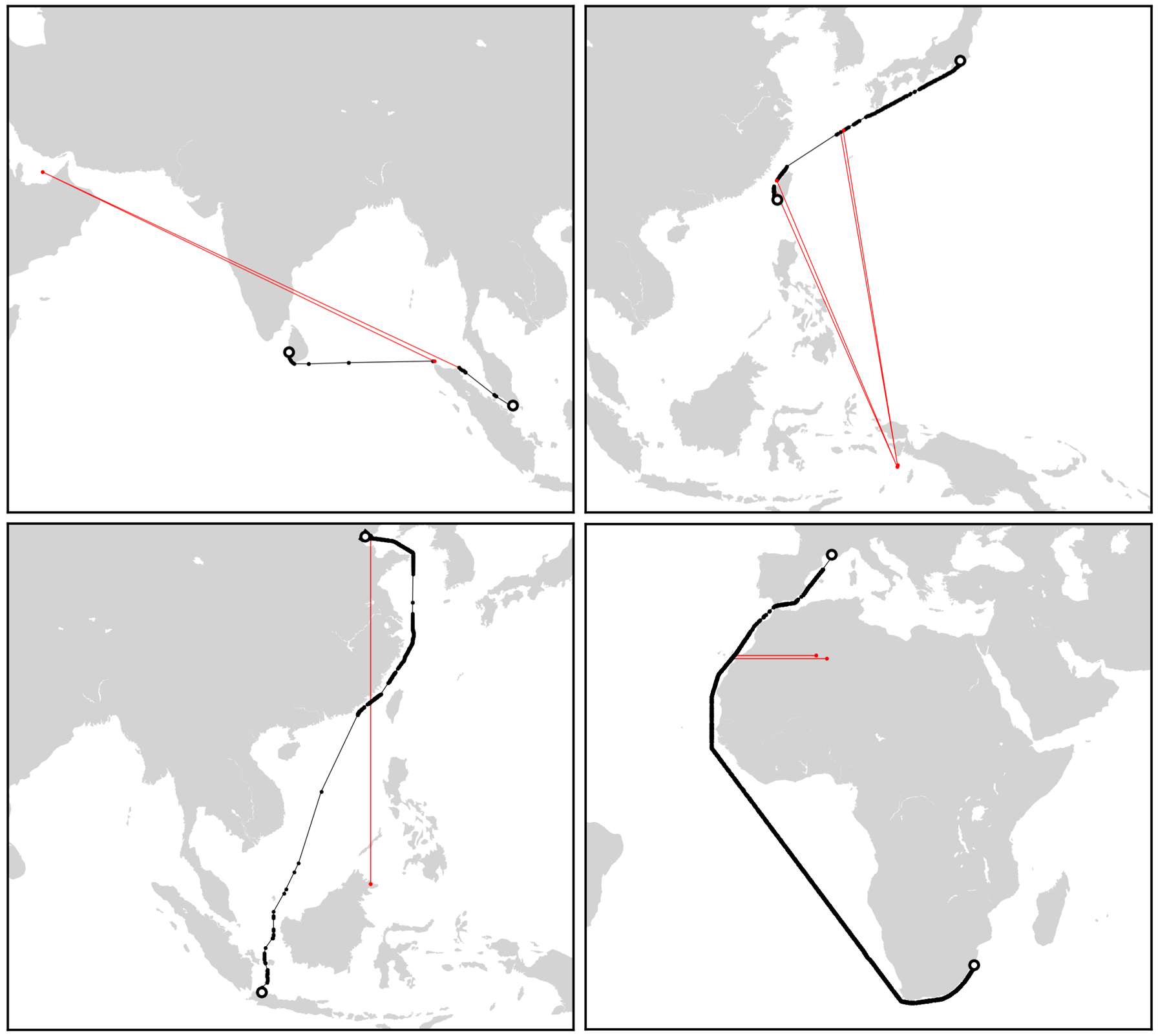}
\caption{Examples of illogical movement detected using the DBSCAN algorithm. The \textcolor{red}{Red} edges are excluded as anomalies.}
\label{fig:abnormal_edge_detection}
\end{figure}
While the annotation process concentrates on capturing the errors that arise from human resources and clarifies the context of trajectories, the refinement process focuses on rectifying machinery errors that generate illogical movements during trajectory progression. 

There are some default values in the kinetic features included in AIS messages, such as SOG of $1023$, COG of $360$, ROT of $-731$, and $511$ heading.
The values indicate the definite error signs reasoned by the out-of-field range; therefore, the removal must be performed.
However, in addition to the clear signs, illogical shifts still exists in the annotated trajectories.
These anomalies can occur either by the combination or by the sole of unspecified transmission errors.

The density-based clustering algorithm, DBSCAN, is utilized to capture edge-wise anomalies derived from unrecognized errors.
Using the time, spatial distances, and SOGs between each continuous observation pair, residual vectors are composed to represent the normality of edges, where each vector is given as follows: $\{\text{ } |\lambda_\Delta|, |\phi_\Delta|, ||{\lambda_\Delta, \phi_\Delta}|| \text{ }\} / (|\delta_{time}|\odot \text{AVG}(\mathrm{SOG})) $, where the residual of each longitude and latitude are $\lambda_\Delta$ and $\phi_\Delta$, $|\cdot|$ denotes the Manhattan distance, $||\cdot||$ is Euclidean, $\delta_{time}$ indicates the time interval between two, and $\text{AVG}(\cdot)$ is the average.
The hyperparameter of the DBSCAN algorithm, epsilon, is set to 0.15, and the results of abnormal detection from the edge-wise vectors are shown in Fig \ref{fig:abnormal_edge_detection}.
By deleting these illogical conclusions, the trajectories can possibly lose their validity if the arrival is betided by such errors.
Consequently, the sequences must be re-validated by the last sub-procedure defined in Appendix \ref{appendix:port_label_annotation}.

\end{document}